%% file: main.tex
\documentclass[journal]{IEEEtran}
\usepackage{amsmath,amssymb,amsfonts}
\usepackage{graphicx}
\usepackage{textcomp}

\usepackage{amsthm}
\usepackage{xcolor}
\usepackage[figuresright]{rotating}

\usepackage{graphicx}
\graphicspath{{/}{fig/}}

\usepackage{array}
\usepackage{textcomp}
\usepackage{xcolor}
\usepackage{multirow}
\usepackage{booktabs}
\usepackage{enumitem}

\usepackage{mathtools}
\usepackage{breqn}
\usepackage{float}

\usepackage{caption}
\usepackage{subcaption}%for subfigures as well

\usepackage{mathtools}
\usepackage{float}
\usepackage{hyperref}

\usepackage{pgfplots}
\pgfplotsset{compat=1.7}
\usepgfplotslibrary{groupplots}

\usepackage{multirow,tabularx}

\newlength\figureheight
\newlength\figurewidth
\title{\LARGE \bf
    Ubiquitous Distributed Deep Reinforcement Learning at the Edge: Analyzing Byzantine Agents in Discrete Action Spaces
}

\author{
    \IEEEauthorblockN{
        Wenshuai Zhao\textsuperscript{1},
        Jorge Pe\~{n}a Queralta\textsuperscript{1},
        Li Qingqing\textsuperscript{1},
        Tomi Westerlund\textsuperscript{1}
    }\\[6pt]
    \IEEEauthorblockA{
        \textsuperscript{1} \href{https://tiers.utu.fi}{Turku Intelligent Embedded and Robotic Systems Lab, University of Turku, Finland} \\
        Emails: \textsuperscript{1}\{wezhao, jopequ, qingqli, tovewe\}@utu.fi
    }
}

\begin{document}

%%%%%%%%%%%%%%%%%%%%%%%%%%%%%%%%%%%%%%%%%%%%%%
%%                                          %%
%%              PAPER CONTENT               %%
%%                                          %%
%%%%%%%%%%%%%%%%%%%%%%%%%%%%%%%%%%%%%%%%%%%%%%
\maketitle

\input{sections/00_Abstract.tex}
\input{sections/01_Introduction.tex}
\input{sections/02_RelatedWorks}
\input{sections/03_Methodology}
\input{sections/04_ExperimentsAndResults.tex}   
\input{sections/05_Conclusion.tex}

%%%%%%%%%%%%%%%%%%%%%%%%%%%%%%%%%%%%%%%%%%%%%%
%%                                          %%
%%     ACKNOWLEDGEMENT AND BIBLIOGRAPHY     %%
%%                                          %%
%%%%%%%%%%%%%%%%%%%%%%%%%%%%%%%%%%%%%%%%%%%%%%

\section*{Acknowledgements}

This work was supported by the Academy of Finland's AutoSOS project with grant number 328755.

% \nocite{*}
\bibliographystyle{unsrt}
\bibliography{ref}

% \clearpage
% \newpage
% \Large{\textbf{Authors' Background}} \\[+23pt]
% \normalsize
% \begin{tabular}{@{}llll@{}}
%     \toprule
%     \textbf{Name} & \textbf{Title} & \textbf{Research Field} & \textbf{Personal Website} \\
%     \midrule
%     Wenshuai Zhao & Master student & Embedded and Robotic Systems &     tiers.utu.fi \\
%     Jorge Peña Queralta & PhD Student & Embedded and Robotic Systems & tiers.utu.fi \\
%     Qingqing Li & PhD Student & Embedded and Robotic Systems & tiers.utu.fi \\
%     Tomi Westerlund & Associate Professor & Embedded and Robotic Systems & tiers.utu.fi \\
%     \bottomrule
% \end{tabular}

\end{document}

%% file: sections/00_Abstract.tex
\begin{abstract}

    The integration of edge computing in next-generation mobile networks is bringing low-latency and high-bandwidth ubiquitous connectivity to a myriad of cyber-physical systems. This will further boost the increasing intelligence that is being embedded at the edge in various types of autonomous systems, where collaborative machine learning has the potential to play a significant role. This paper discusses some of the challenges in multi-agent distributed deep reinforcement learning that can occur in the presence of byzantine or malfunctioning agents. As the simulation-to-reality gap gets bridged, the probability of malfunctions or errors must be taken into account. We show how wrong discrete actions can significantly affect the collaborative learning effort. In particular, we analyze the effect of having a fraction of agents that might perform the wrong action with a given probability. We study the ability of the system to converge towards a common working policy through the collaborative learning process based on the number of experiences from each of the agents to be aggregated for each policy update, together with the fraction of wrong actions from agents experiencing malfunctions. Our experiments are carried out in a simulation environment using the Atari testbed for the discrete action spaces, and advantage actor-critic (A2C) for the distributed multi-agent training.

\end{abstract}

\begin{IEEEkeywords}
    Reinforcement Learning; Edge Computing; Multi-Agent Systems; Collaborative Learning; RL; Deep RL; Adversarial RL;
\end{IEEEkeywords}

\IEEEpeerreviewmaketitle

%% file: sections/01_Introduction.tex
\section{Introduction}

The edge computing paradigm is bringing higher degrees of intelligence to connected cyber-physical systems across multiple domains. This intelligence is being in turn enabled by lightweight deep learning (DL) models deployed at the edge for real-time computation. Among the multiple DL approaches, reinforcement learning (RL) has been increasingly adopted in various types of cyber-physical systems over the past decade, and, in particular, multi-agent RL~\cite{nguyen2020deep}. Deep reinforcement learning (DRL) algorithms are motivated by the way natural learning happens: through trial and error, learning from experiences based on the performance outcome of different actions. Among other fields, DRL algorithms have had success in robotic manipulation~\cite{matas2018sim}, but also in finding more optimal approaches to complex multi-dimensional problems involved in edge computing~\cite{ning2019deep}.

We are particularly interested on how edge computing can enable more efficient real-time distributed and collaborative multi-agent RL. When discussing multi-agent DRL, two different views emerge from the literature: those where multiple agents are utilized to improve the learning process (e.g., faster learning through parallelization~\cite{mnih2016asynchronous}, higher diversity by means of exploration of different environments~\cite{raileanu2018modeling}, or increased robustness with redundancy~\cite{nguyen2020deep}), and those where multiple agents are learning a policy emerging from an interactive behavior (e.g., formation control algorithms~\cite{conde2017time}, or collision avoidance~\cite{long2018towards}).

\begin{figure*}
    \centering
    \includegraphics[width=\textwidth]{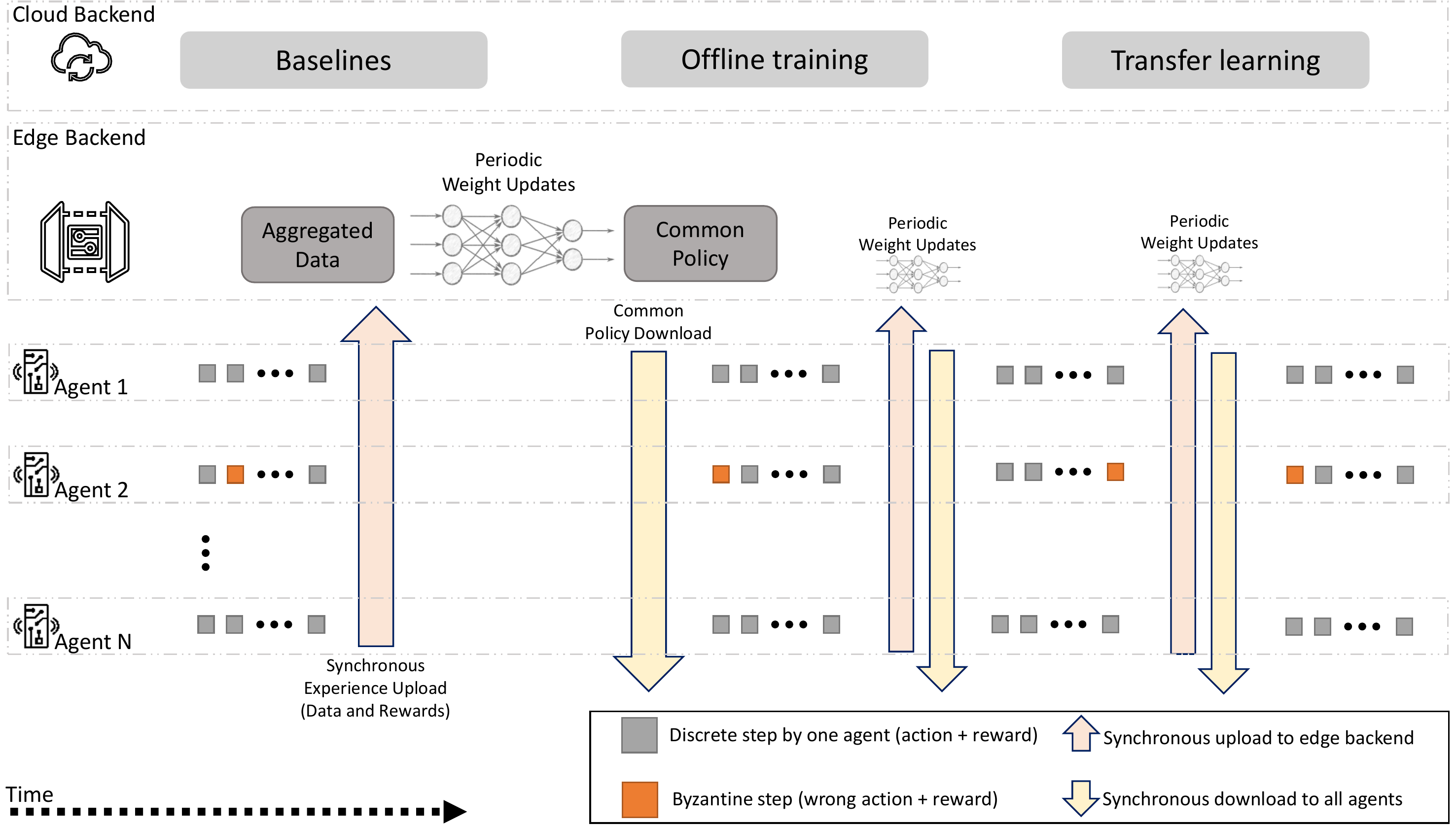}
    \caption{Conceptual view of the system architecture proposed in this paper. Multiple agents are collaborating towards learning a common task, with the deep neural network updates being synchronized at the edge layer. Each agent, however, individually explores its environments gathering experience in a series of episodes, and calculating the corresponding rewards.}
    \vspace{-1em}
    \label{fig:concept}
\end{figure*}

This work explores an scenario where multiple agents are collaboratively learning towards the same task, which is often individual, as illustrated in Fig.~\ref{fig:concept}. Deep reinforcement learning has been identified as one of the key areas that will see wider adoption within the Internet of Things (IoT) owing to the rise of edge computing and 5G-and-beyond connectivity~\cite{queralta2020enhancing, queralta2020blockchain}. Multiple application fields can benefit from this synergy in different domains, from robotic agents in the industrial IoT, to information fusion in wireless sensor networks, and including all types of connected autonomous systems and other types of cyber-physical systems.

Multiple challenges still exist in distributed multi-agent DRL. Among the most relevant ones within the scope of this paper are the development of novel techniques to increase robustness in the presence of adversarial agents or perturbed environments~\cite{behzadan2017vulnerability, gleave2019adversarial, wang2020reinforcement}, as well as closing the simulation-to-reality gap~\cite{matas2018sim, balaji2019deepracer, wenshuai2020sim2real}. In relation to the former area, recent works have focused on exploring different types of noise or perturbations in the agents or environments to better understand how these potentially adversarial conditions affect the collaborative learning process. For example, Gu et al. study in~\cite{gu2017deep} the effect of network delays and propose an asynchronous method for off-policy updates, while Yu et al. have studied the effect of adversarial conditions in the network connection between the agents~\cite{yu2020multi}. We have seen a lack of research, however, on the analysis of adversarial conditions in discrete action spaces. These type of scenarios occur when agents need to make a decision from a finite and discrete set of actions.

In this paper, we study the effect of byzantine agents that perform the wrong action with certain probabilities and report initial results that let us understand the limitations of the state-of-the-art in multi-agent RL in the presence of byzantine agents for discrete action spaces. In particular, we utilize the synchronous advantage actor-critic (A2C) algorithm on two of the standard Atari environments typically used for benchmarking DRL methods. This is, to the best of our knowledge, the first paper analyzing the effects terms of policy convergence in collaborative multi-agent DRL caused by having different fractions of wrong actions in discrete action spaces. Our results show that in some environments with totally 16 distributed agents the training process is highly sensitive to having a single agent acting in the wrong manner over a relatively small fraction of its actions, and unstable convergence appears with just a single agent having 2.5\% of its actions wrong. In other environments the threshold is higher, with the network still converging to a working policy at over 10\% of wrong actions in a single byzantine agent.

The remainder of this document is organized as follows. Section 2 presents related works in the area of adversarial RL and applications combining RL with edge computing. Section 3 describes the basic theory behind A2C and the simulation environments. Section 4 then presents our results on the convergence of the system when a fraction of the actions is wrong, and Section 5 concludes the work and outlines our future work directions.
% \vspace{-2em}

%% file: sections/02_RelatedWorks.tex
\section{Related Works}
Adversarial RL has attracted many researchers' interest in recent years. Multiple deep learning algorithms are known to be vulnerable to manipulation by perturbed inputs~\cite{behzadan2017vulnerability}.  This problem also affects various reinforcement learning algorithms under different scenarios. In multi-agent environments, an attacker can significantly increase the adversarial observations ability~\cite{gleave2019adversarial}. Ilahi et al. review emerging adversarial attacks in DRL-based systems and the potential countermeasures to defend against these attacks ~\cite{ilahi2020challenges}.  The authors classify the attacks as attacks targeting (i) rewards, (ii) policies, (iii) observations, and (iv) the environment. In this paper, instead, we consider targeting the agents' actions, which can happen in real-world applications when agents interact with their environment.

Similarly considering how to better transfer learning from simulations to the real-world, multiple researchers have been working on a simulation-to-reality transfer for specific applications in different environments~\cite{balaji2019deepracer, matas2018sim, arndt2019meta}. In this paper, we will analyze the effect of the adversarial of byzantine effects in multi-agents reinforcement learning, and introduce a fraction of byzantine actions, which has not been studied before. 

Other researchers have explored the influence of noisy rewards in RL. Wang et al. present a robust RL framework that enables agents to learn in noisy environments where only perturbed rewards are observed, and analyzes different algorithms performance under their proposed framework, including PPO, DQN, and DDPG~\cite{wang2020reinforcement}. In this paper, the perturbances on the DRL process will be explored, but we focus on analyzing discrete action spaces and a fraction of byzantine actions performed by a small number of byzantine agents.
%Also, we put more effort into multiple agents' cooperation learning in a different perturbances situation. We mainly explore the PPO algorithm which is the state-of-art algorithms in three-dimensional locomotion. PPO algorithm has shown significant robust performance against reward perturbances ~\cite{wang2020reinforcement} and noisy rewards in ~\cite{kumar2019enhancing}. 

%\blue{Other authors have explored the effects of having noisy rewards in RL. In this direction, Wang et al. presented an analysis of perturbed rewards for different RL algorithms, including PPO but also DQN and DDPG, among others~\cite{wang2020reinforcement}.Compared to their approach, we also consider perturbances on the RL process but focus on those that model real-world noises and errors. 
%\{Moreover, we specifically put an emphasis on multi-robot collaborative learning, and consider situations in which the perturbances that affect different robots are also different. We also focus on the PPO algorithm as the state-of-the-art in three-dimensional locomotion. In fact, PPO has been identified as one of the most robust approaches against reward perturbances in~\cite{wang2020reinforcement}. Also within the study of noisy rewards, a method to improve performance in such scenarios is proposed in~\cite{kumar2019enhancing}.}

Multiple works have been presented in the convergence of DRL and edge computing. However, rather than focusing on exploiting edge computing for distributed RL, most of the current literature is exploiting RL for edge service. For instance, Ning et al. apply DRL for more efficient offloading orchestration~\cite{ning2019deep}, while Wang et al. have applied DRL to optimize resource allocation at the edge~\cite{wang2019smart}. In our work, however, we focus on analyzing some of the challenges that can appear when edge computing is exploited for distributed multi-agent DRL in real-world applications.

%% file: sections/03_Methodology.tex
\section{Methodology}

This section describes the methods and simulation environments utilized for the analysis in Section 4: the advantage actor-critic (A2C) algorithm for distributed DRL, and the simulation environment.

Actor-critic methods combine the advantages of value based and policy based methods, and has been regarded as the base of many modern RL algorithms. In A2C, two neural networks represent the actor and critic, where the actor controls the agent's behavior and the critic evaluate how good the action taken is. As value-based methods tend to high variability, an advantage function is employed to replace the raw value function, leading to advantage actor-critic (A2C). The main scheme of the policy gradient updates is shown in \eqref{eq:update}:
\begin{equation}
    \theta^{new}\leftarrow\\\theta^{old}+\eta\nabla{{\overline{R}_{\theta}}}
    \label{eq:update}
\end{equation}
where $\theta$ denotes the policy to be learned, $\eta$ is the learning rate, and $\nabla{{\overline{R}_{\theta}}}$ represents the policy gradient, given by \eqref{eq:gradient}.
\begin{equation}
    \nabla{\overline{R}_{\theta}}\approx\\\dfrac{1}{N}\sum_{n=1}^{N}\sum_{t=1}^{T_{n}}\\R(\tau^{n})\nabla\\logp(a_{t}^{n}\vert\\s_{t}^{n},\theta)
    \label{eq:gradient}
\end{equation} 
where $N$ is the number of trajectories sampled under the policy $\tau$, and $R(\tau^{n})$ denotes the accumulated reward for each episode consisting of $T_{n}$ steps. In the policy with weight $\theta$, an action $a_{t}^{n}$ under the state $s_{t}^{n}$ is chosen with probability $p(a_{t}^{n}\vert s_{t}^{n},\theta)$. %in the policy $\tau$ 

In this policy gradient method, the accumulated reward $R(\tau^{n})$ is calculated by sampling the trajectories, which are computed when the whole episode is finished, and hence might bring high variability affecting the policy convergence. To avoid this, value estimation is introduced and merged into the policy gradient method. An advantage function is thus proposed to replace $R(\tau^{n})$ according to \eqref{eq:advantage}, which is also the reason for the name of A2C.
\begin{equation}
    R(\tau^{n})=r_{t}^{n}+V^{\pi}(s_{t+1}^{n})-V^{\pi}(s_{t}^{n})
    \label{eq:advantage}
\end{equation} 
where $r_{t}$ is the reward gained in the step $t$, and $V^{\pi}$ denotes the value function to estimate the accumulated reward that will be gained. Additionally, in the implementation of this A2C algorithm, multiple agents are employed to produce the trajectories in parallel. Compared to A3C~\cite{mnih2016asynchronous}, in which each agent will update the network individually and asynchronously, A2C collects the whole data from each agents and then update the shared network. This is also illustrated in Fig.~\ref{fig:concept}.

In order to analyze the effect of byzantine actions, we choose two typical gym-wrapped Atari games as our simulation environments: PongNoFrameskip-v4 (Fig.~\ref{fig:pong_env}) and BreakoutNoFrameskip-v4 (Fig.~\ref{fig:breakout_env}). Both of them take video as an input, based on which the policy will be trained to produce the corresponding discrete actions to obtain higher rewards. The action spaces for Pong and Breakout have cardinality of 5 and 4, respectively. We set their corresponding byzantine agents to behave with the opposite actions (e.g., if the output action from the policy is $action=2$ in Pong, then a wrong action by the byzantine agents will be $action=3$). In this paper, we consider the effect of the presence of Byzantine agent in terms of their number and the frequency of wrong actions they perform. The patterns we observe in the experiments can be further utilized to detect Byzantine agents in distributed multi-agent DRL scenarios. 

%% file: sections/04_ExperimentsAndResults.tex
\begin{figure*}
    \centering
    \begin{subfigure}{0.7\textwidth}
        \centering
        \setlength{\figureheight}{0.5\textwidth}
        \setlength{\figurewidth}{\textwidth}
        \small{\input{fig/pong/logdir_0B}}
        \caption{Reference training (no byzantine agents).}
        \label{fig:pong_logdir_0B}
    \end{subfigure}
    \begin{subfigure}{0.27\textwidth}
        \centering
        \includegraphics[width=.95\textwidth]{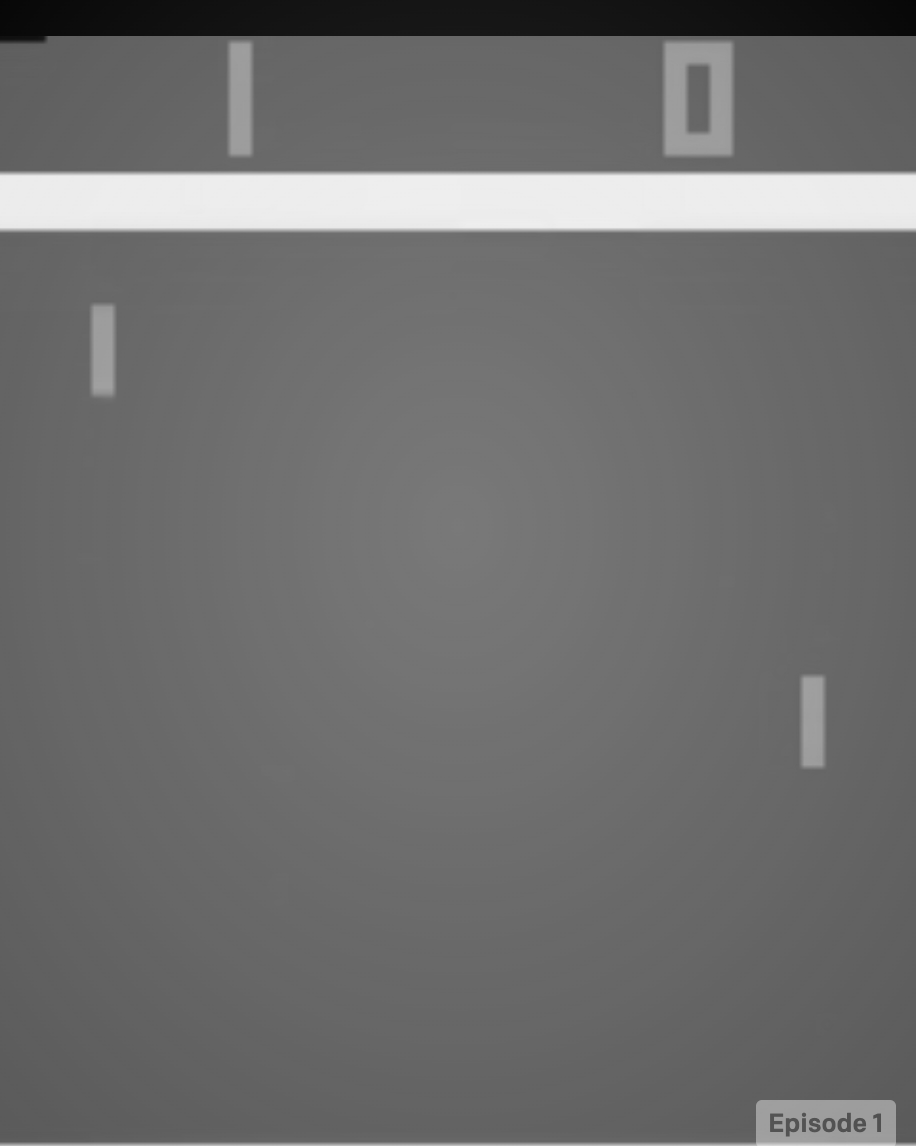}
        \caption{PongNoFrameskip-v4 environment.}
        \label{fig:pong_env}
    \end{subfigure}
    %Added by Wenshuai
    \begin{subfigure}{0.32\textwidth}
        \centering
        \setlength{\figureheight}{0.6\textwidth}
        \setlength{\figurewidth}{\textwidth}
        \scriptsize{\input{fig/pong/logdir_1B}}
        \caption{One agent with continuous byzantine actions.}
        \label{fig:pong_logdir_1B}
    \end{subfigure}
    % \begin{subfigure}{0.32\textwidth}
    %     \centering
    %     % \includegraphics[width=\textwidth]{fig/pong/logdir_0B.pdf}
    %     \setlength{\figureheight}{0.6\textwidth}
    %     \setlength{\figurewidth}{\textwidth}
    %     \scriptsize{\input{fig/pong/logdir_2B}}
    %     \caption{Two Continuous Byzantine agents}
    %     \label{fig:pong_logdir_2B}
    % \end{subfigure}
    %END
    \begin{subfigure}{0.32\textwidth}
        \centering
        \setlength{\figureheight}{0.6\textwidth}
        \setlength{\figurewidth}{\textwidth}
        \scriptsize{\input{fig/pong/logdir_1B_5m1_p}}
        \caption{One agent, byzantine actions in 1/5 updates (20\% of the total).}
        \label{fig:pong_logdir_1B_5m1}
    \end{subfigure}
    \begin{subfigure}{0.32\textwidth}
        \centering
        \setlength{\figureheight}{0.6\textwidth}
        \setlength{\figurewidth}{\textwidth}
        \scriptsize{\input{fig/pong/logdir_1B_10m1_p}}
        \caption{One agent, byzantine actions in 1/10 updates (10\% of the total).}
        \label{fig:pong_logdir_1B_10m1}
    \end{subfigure}
    \begin{subfigure}{0.32\textwidth}
        \centering
        \setlength{\figureheight}{0.6\textwidth}
        \setlength{\figurewidth}{\textwidth}
        \scriptsize{\input{fig/pong/logdir_1B_10m2_p}}
        \caption{One agent, byzantine actions in 1/2 steps of 1/10 updates (5\% of the total).}
        \label{fig:pong_logdir_1B_10m2}
    \end{subfigure}
    \begin{subfigure}{0.32\textwidth}
        \centering
        \setlength{\figureheight}{0.6\textwidth}
        \setlength{\figurewidth}{\textwidth}
        \scriptsize{\input{fig/pong/logdir_2B_10m2_p}}
        \caption{Two agents, byzantine actions in 1/2 steps of 1/10 updates ($2\times5$\% of the total).}
        \label{fig:pong_logdir_2B_10m2}
    \end{subfigure}
    \begin{subfigure}{0.32\textwidth}
        \centering
        \setlength{\figureheight}{0.6\textwidth}
        \setlength{\figurewidth}{\textwidth}
        \scriptsize{\input{fig/pong/logdir_4B_20m2_p}}
        \caption{Four agents, byzantine actions in 1/2 steps of 1/20 updates ($4\times2.5$\% of the total).}
        \label{fig:pong_logdir_4B_20m2}
    \end{subfigure}
    \caption{Experiments in the PongNoFrameskip-v4 environment.}
    \label{fig:pong_resulst}
\end{figure*}

\section{Simulations and Results}

\begin{figure*}
    \centering
    \begin{subfigure}{0.7\textwidth}
        \centering
        \setlength{\figureheight}{0.4\textwidth}
        \setlength{\figurewidth}{\textwidth}
        \small{\input{fig/breakout/logdir_0B_0713}}
        \caption{Reference training (no byzantine agents)}
        \label{fig:breakout_logdir_0B}
    \end{subfigure}
    \begin{subfigure}{0.27\textwidth}
        \centering
        \includegraphics[width=.75\textwidth]{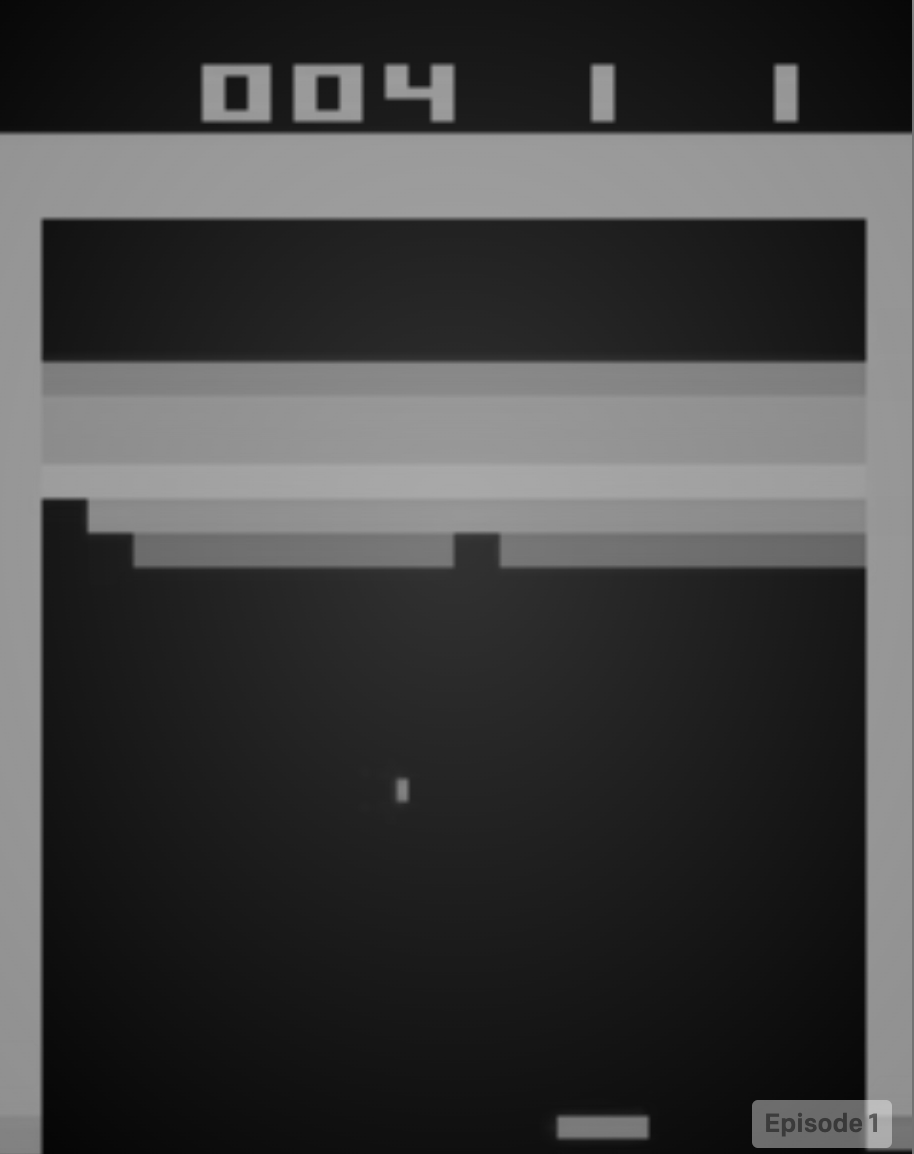}
        \caption{BreakoutNoframeskip-v4 env.}
        \label{fig:breakout_env}
    \end{subfigure}

    \begin{subfigure}{0.32\textwidth}
        \centering
        \setlength{\figureheight}{0.6\textwidth}
        \setlength{\figurewidth}{\textwidth}
        \scriptsize{\input{fig/breakout/logdir_1B_10m1_p}}
        \caption{One agent, byzantine actions in 1/10 updates (10\% of the total).}
        \label{fig:breakout_1B_10m1}
    \end{subfigure}
    \begin{subfigure}{0.32\textwidth}
        \centering
        \setlength{\figureheight}{0.6\textwidth}
        \setlength{\figurewidth}{\textwidth}
        \scriptsize{\input{fig/breakout/logdir_1B_10m2_p}}
        \caption{One agent, byzantine actions in 1/2 steps of 1/10 updates (5\% of the total).}
        \label{fig:breakout_1B_10m2}
    \end{subfigure}
    \begin{subfigure}{0.32\textwidth}
        \centering
        \setlength{\figureheight}{0.6\textwidth}
        \setlength{\figurewidth}{\textwidth}
        \scriptsize{\input{fig/breakout/logdir_1B_20m2_p}}
        \caption{One agent, byzantine actions in 1/2 steps of 1/20 updates (2.5\% of the total).}
        \label{fig:breakout_1B_20m2}
    \end{subfigure}
    \begin{subfigure}{0.32\textwidth}
        \centering
        \setlength{\figureheight}{0.6\textwidth}
        \setlength{\figurewidth}{\textwidth}
        \scriptsize{\input{fig/breakout/logdir_1B_5m2_p}}
        \caption{One agent, byzantine actions in 1/2 steps of 1/5 updates (10\% of the total).}
        \label{fig:breakout_1B_5m2}
    \end{subfigure}
    \begin{subfigure}{0.32\textwidth}
        \centering
        \setlength{\figureheight}{0.6\textwidth}
        \setlength{\figurewidth}{\textwidth}
        \scriptsize{\input{fig/breakout/logdir_2B_10m2_p}}
        \caption{Two agents, byzantine actions in 1/2 steps of 1/10 updates ($2\times5$\% of the total).}
        \label{fig:breakout_4B_20m2}
    \end{subfigure}
    \begin{subfigure}{0.32\textwidth}
        \centering
        \setlength{\figureheight}{0.6\textwidth}
        \setlength{\figurewidth}{\textwidth}
        \scriptsize{\input{fig/breakout/logdir_4B_20m2_p}}
        \caption{Four agents, byzantine actions in 1/2 steps of 1/20 updates ($4\times2.5$\% of the total).}
        \label{fig:breakout_4B_20m2}
    \end{subfigure}
    % \begin{subfigure}{0.32\textwidth}
    %     \centering
    %     \setlength{\figureheight}{0.6\textwidth}
    %     \setlength{\figurewidth}{\textwidth}
    %     \scriptsize{\input{fig/breakout/logdir_1B}}
    %     \caption{4B\_20m2}
    %     \label{fig:breakout_4B_20m2}
    % \end{subfigure}
    \caption{Experiments in the BreakoutNoFrameskip-v4 environment.}
    \vspace{-1em}
    \label{fig:breakout_results}
\end{figure*}

In this section, we describe the settings in our experiments and present the main conclusions of our analysis. There are totally 16 agents or workers employed to produce trajectory data %and similar settings are deployed 
both in Pong and Breakout environments. The experiences, actions and rewards from the different agents are then aggregated synchronously to calculate the policy gradients and update the policy towards a more optimal one. In the experiments, we analyze how byzantine agents that perform wrong actions unknowingly affect the collaborative learning effort. The reference training without byzantine agents for the Pong and Breakout environments are shown in Fig.~\ref{fig:pong_logdir_0B} and Fig.~\ref{fig:breakout_logdir_0B}, respectively.

%are then used to calculate the policy gradients and obtain the optimal policy. However, if we set Byzantine ones among these 16 agents, the collected data would be corrupted and hence bring fluctuation to the policy gradients, from which we will observe the differences as we change the deployment of Byzantine agents in order to find some patterns to help detect them.

In the Pong environment, we first set a single agent continuously behaving wrongly, out of the total of 16 agents working in parallel (Fig.~\ref{fig:pong_logdir_1B}). Compared to the reference training, we observe that the policy is unable to improve in order to obtain better rewards. Therefore, a single byzantine agent representing as little as 6.25\% of the total is enough to completely disable the ability of the system to converge towards a working policy. Therefore, we have focused on analyzing the maximum fraction of wrong actions that byzantine agents can perform in order to ensure convergence of the system. Moreover, in order to test whether it is the total fraction what matters or the number of agents, we have considered the same total fraction of byzantine actions in different settings.

In the training, the policy is updated only when the agents perform a series of steps, collecting a certain amount of interaction data. In particular, agents perform 5 steps between updates of the policy. This leads to 80 steps between updates, and we set the total number of steps to $10^7$ for the complete training process. The number of episodes depends on the performance of the agents (the better the performance, the longer the episodes are).

With this, we set different fractions of byzantine actions depending on (i) the number of agents, (ii) the number of wrong actions in between updates, and (iii) the fraction of updates affected by byzantine actions. The results for the Pong environment are shown in Figures~\ref{fig:pong_logdir_1B_5m1} through~\ref{fig:pong_logdir_4B_20m2}. From these, we conclude that $20\%$ of byzantine actions are enough to deplete the systems' ability to converge, while the system is able to converge with slight unstabilities in the presence of a 10\% of byzantine actions (Figures~\ref{fig:pong_logdir_1B_10m1}, \ref{fig:pong_logdir_2B_10m2} and~\ref{fig:pong_logdir_4B_20m2}).  Finally, with just 5\% of byzantine actions (Fig.~\ref{fig:pong_logdir_1B_10m2}) the convergence is similar to the reference.

%By the comparison of convergence between normal training as Figure~\ref{fig:pong_logdir_0B}, and one Byzantine agent as Figure~\ref{fig:pong_logdir_1B} in the total 16 agents, we find that only one out of 16 continuous behaving Byzantine agent will disastrously destroy the training convergence. Therefore, we further fine tune the frequency of Byzantine actions. In the training, the policy is updated the only when collecting certain interaction data, and for every collecting, each agent will repeat five steps to sample. In this way, we set different intervals of Byzantine actions in both the update level and repeated step level. The notation describing such settings is $bB\_ums$, in which $b$ represents the number of Byzantine agents in the whole 16 agents, $u$ denotes the interval between updates when we set Byzantine actions, and $s$ means the interval between repeated Byzantine step in each Byzantine update. The results are shown as Figure~\ref{fig:pong_logdir_1B_5m1}, Figure~\ref{fig:pong_logdir_1B_10m1}, Figure~\ref{fig:pong_logdir_1B_10m2}, Figure~\ref{fig:pong_logdir_2B_10m2}, and Figure~\ref{fig:pong_logdir_4B_20m2}, from which it is easy to find that $20\%$ Byzantine actions lead to bad convergence while the total $10\%$ Byzantine actions still could get well converged, no matter the Byzantine actions behave in the update level or repeated step level.

In addition, we also conduct similar experiments on another classical Atari game, BreakoutNoFrameskip-v4. The results with byzantine actions are shown in Figures~\ref{fig:breakout_1B_10m1} through Figure~\ref{fig:breakout_4B_20m2}.  In this environment, $10\%$ byzantine actions are enough to deter convergence (Fig.~\ref{fig:breakout_1B_10m1}, Figure~\ref{fig:breakout_1B_5m2}, and Figure~\ref{fig:breakout_4B_20m2}). Only by reducing the frequency to $2.5\%$ can the training get acceptable convergence (Figure~\ref{fig:breakout_1B_20m2}). A general conclusion is also that the total fraction of byzantine actions is what matters the most, and not how they are introduced in the system.

%We have seen that the policy converged unstably in one of the environments with just 1/40 byzantine agents performed by a single byzantine agent (out of 16)

%% file: fig/pong/logdir_0B.tex
% This file was created by tikzplotlib v0.9.2.
\begin{tikzpicture}

% \begin{axis}[
% axis background/.style={fill=white!82.7450980392157!black},
% axis line style={white},
% legend cell align={left},
% legend style={fill opacity=0.8, draw opacity=1, text opacity=1, at={(1,0.1)}, anchor=south east, draw=white!80!black},
% tick align=outside,
% tick pos=left,
% x grid style={white!69.0196078431373!black},
% xlabel={Episode},
% xmajorgrids,
% xmin=-15.85, xmax=354.85,
% xtick style={color=black},
% y grid style={white!69.0196078431373!black},
% ylabel={Reward},
% ymajorgrids,
% ymin=-23.1, ymax=23.1,
% ytick style={color=black}
% ]
\definecolor{color0}{rgb}{0.83921568627451,0.152941176470588,0.156862745098039}
\definecolor{color1}{rgb}{0.12156862745098,0.966666666666667,0.205882352941177}
\definecolor{color2}{rgb}{0.580392156862745,0.503921568627451,0.941176470588235}
\definecolor{color3}{rgb}{0.290196078431372,0.166666666666667,0.23078431372549}

\begin{axis}[
    width=\figurewidth,
    height=\figureheight,
    axis background/.style={fill=white},
    axis line style={white},
    legend cell align={left},
    legend style={fill opacity=0.8, draw opacity=1, text opacity=1, at={(1,0.1)}, anchor=south east, draw=white!80!black},
    tick align=outside,
    tick pos=left,
    x grid style={white},
    xlabel={Episode},
    ymin=-25,
    ymax=25,
    xmajorgrids,
    xtick style={color=black},
    y grid style={white!80!black},
    ylabel={Reward},
    ymajorgrids,
    ytick style={color=black},
    scaled y ticks = false,
    scaled x ticks = false
]
\addplot [semithick, color2, opacity=0.5]
table {%
1 -20
2 -21
3 -19
4 -21
5 -19
6 -20
7 -20
8 -21
9 -21
10 -20
11 -20
12 -21
13 -21
14 -20
15 -21
16 -21
17 -21
18 -21
19 -21
20 -21
21 -21
22 -21
23 -20
24 -20
25 -21
26 -21
27 -21
28 -21
29 -20
30 -21
31 -21
32 -20
33 -21
34 -21
35 -20
36 -21
37 -21
38 -21
39 -20
40 -19
41 -21
42 -20
43 -20
44 -20
45 -21
46 -19
47 -20
48 -20
49 -21
50 -21
51 -18
52 -20
53 -20
54 -19
55 -17
56 -20
57 -20
58 -20
59 -21
60 -20
61 -18
62 -16
63 -14
64 -20
65 -13
66 -11
67 -16
68 -7
69 -18
70 -16
71 -5
72 -14
73 -12
74 -4
75 3
76 1
77 -2
78 1
79 6
80 8
81 8
82 4
83 8
84 7
85 10
86 7
87 10
88 13
89 17
90 14
91 9
92 18
93 18
94 17
95 20
96 18
97 17
98 18
99 18
100 18
101 17
102 19
103 19
104 18
105 19
106 19
107 16
108 20
109 19
110 -14
111 7
112 20
113 17
114 -1
115 -7
116 20
117 14
118 18
119 19
120 19
121 18
122 18
123 20
124 20
125 19
126 17
127 19
128 19
129 18
130 21
131 -7
132 -21
133 -21
134 13
135 16
136 19
137 19
138 19
139 11
140 18
141 18
142 17
143 17
144 16
145 20
146 17
147 17
148 19
149 19
150 20
151 16
152 17
153 -20
154 -21
155 -21
156 -21
157 -19
158 16
159 15
160 19
161 18
162 17
163 19
164 17
165 19
166 20
167 18
168 19
169 17
170 19
171 18
172 20
173 18
174 18
175 20
176 18
177 21
178 20
179 18
180 15
181 15
182 15
183 18
184 19
185 19
186 19
187 18
188 21
189 19
190 20
191 18
192 18
193 19
194 19
195 20
196 19
197 21
198 20
199 20
200 18
201 17
202 20
203 18
204 17
205 20
206 19
207 19
208 20
209 18
210 17
211 -1
212 7
213 20
214 20
215 18
216 17
217 21
218 18
219 20
220 19
221 19
222 3
223 19
224 19
225 17
226 18
227 21
228 18
229 21
230 20
231 20
232 20
233 20
234 16
235 21
236 19
237 20
238 17
239 20
240 19
241 19
242 17
243 17
244 19
245 19
246 21
247 19
248 20
249 19
250 19
251 21
252 20
253 2
254 19
255 19
256 20
257 19
258 18
259 21
260 19
261 18
262 17
263 17
264 20
265 21
266 20
267 21
268 21
269 21
270 20
271 19
272 18
273 19
274 19
275 20
276 19
277 19
278 15
279 21
280 19
281 19
282 19
283 19
284 19
285 18
286 19
287 20
288 21
289 20
290 20
291 19
292 19
293 16
294 19
295 18
296 21
297 20
298 20
299 20
300 21
301 20
302 19
303 18
304 17
305 16
306 20
307 19
308 20
309 20
310 19
311 19
312 20
313 19
314 20
315 19
316 19
317 20
318 21
319 20
320 19
321 20
322 20
323 20
324 19
325 21
326 19
327 19
328 20
329 18
330 20
331 19
332 20
333 20
334 20
335 20
336 21
337 21
338 19
};
%\addlegendentry{Training}
\end{axis}

\end{tikzpicture}

%% file: fig/pong/logdir_1B.tex
% This file was created by tikzplotlib v0.9.2.
\begin{tikzpicture}

% \begin{axis}[
% axis background/.style={fill=white!82.7450980392157!black},
% axis line style={white},
% legend cell align={left},
% legend style={fill opacity=0.8, draw opacity=1, text opacity=1, at={(1,0.1)}, anchor=south east, draw=white!80!black},
% tick align=outside,
% tick pos=left,
% x grid style={white!69.0196078431373!black},
% xlabel={Episode},
% xmajorgrids,
% xmin=-37.8, xmax=815.8,
% xtick style={color=black},
% y grid style={white!69.0196078431373!black},
% ylabel={Reward},
% ymajorgrids,
% ymin=-21.2, ymax=-16.8,
% ytick style={color=black}
% ]
\definecolor{color0}{rgb}{0.83921568627451,0.152941176470588,0.156862745098039}
\definecolor{color1}{rgb}{0.12156862745098,0.966666666666667,0.205882352941177}
\definecolor{color2}{rgb}{0.580392156862745,0.503921568627451,0.941176470588235}
\definecolor{color3}{rgb}{0.290196078431372,0.166666666666667,0.23078431372549}

\begin{axis}[
    width=\figurewidth,
    height=\figureheight,
    axis background/.style={fill=white},
    axis line style={white},
    legend cell align={left},
    legend style={fill opacity=0.8, draw opacity=1, text opacity=1, at={(1,0.1)}, anchor=south east, draw=white!80!black},
    tick align=outside,
    tick pos=left,
    x grid style={white},
    xlabel={Episode},
    ymin=-25,
    ymax=25,
    xmajorgrids,
    xtick style={color=black},
    y grid style={white!80!black},
    ylabel={Reward},
    ymajorgrids,
    ytick style={color=black},
    scaled y ticks = false,
    scaled x ticks = false
]
\addplot [semithick, color2, opacity=0.5]
table {%
1 -21
2 -21
3 -21
4 -21
5 -21
6 -21
7 -21
8 -21
9 -21
10 -20
11 -20
12 -21
13 -21
14 -21
15 -21
16 -20
17 -21
18 -21
19 -21
20 -20
21 -21
22 -21
23 -21
24 -21
25 -20
26 -19
27 -21
28 -21
29 -21
30 -21
31 -20
32 -21
33 -20
34 -21
35 -21
36 -20
37 -21
38 -21
39 -20
40 -20
41 -21
42 -19
43 -21
44 -21
45 -21
46 -21
47 -21
48 -21
49 -21
50 -21
51 -21
52 -21
53 -21
54 -21
55 -21
56 -20
57 -20
58 -21
59 -21
60 -19
61 -21
62 -21
63 -21
64 -21
65 -21
66 -21
67 -21
68 -21
69 -21
70 -21
71 -21
72 -21
73 -21
74 -21
75 -21
76 -21
77 -21
78 -21
79 -21
80 -21
81 -21
82 -21
83 -21
84 -19
85 -21
86 -20
87 -20
88 -21
89 -20
90 -21
91 -20
92 -21
93 -20
94 -20
95 -21
96 -21
97 -20
98 -21
99 -21
100 -21
101 -21
102 -21
103 -21
104 -21
105 -20
106 -21
107 -21
108 -20
109 -21
110 -21
111 -21
112 -21
113 -21
114 -17
115 -18
116 -18
117 -19
118 -18
119 -21
120 -18
121 -20
122 -21
123 -21
124 -20
125 -21
126 -21
127 -21
128 -21
129 -21
130 -21
131 -21
132 -21
133 -21
134 -21
135 -21
136 -21
137 -21
138 -21
139 -21
140 -21
141 -21
142 -21
143 -21
144 -21
145 -21
146 -20
147 -21
148 -21
149 -21
150 -20
151 -20
152 -21
153 -19
154 -21
155 -21
156 -19
157 -18
158 -20
159 -20
160 -21
161 -21
162 -21
163 -21
164 -21
165 -21
166 -21
167 -21
168 -21
169 -21
170 -21
171 -21
172 -21
173 -21
174 -21
175 -21
176 -21
177 -21
178 -21
179 -21
180 -21
181 -21
182 -21
183 -21
184 -21
185 -21
186 -21
187 -21
188 -21
189 -21
190 -21
191 -21
192 -21
193 -21
194 -20
195 -21
196 -21
197 -21
198 -21
199 -21
200 -20
201 -21
202 -18
203 -18
204 -20
205 -20
206 -21
207 -21
208 -21
209 -21
210 -21
211 -21
212 -21
213 -21
214 -21
215 -21
216 -21
217 -21
218 -21
219 -21
220 -21
221 -20
222 -21
223 -21
224 -21
225 -21
226 -19
227 -21
228 -20
229 -20
230 -21
231 -21
232 -21
233 -21
234 -21
235 -21
236 -21
237 -21
238 -21
239 -21
240 -21
241 -21
242 -21
243 -21
244 -21
245 -21
246 -21
247 -21
248 -21
249 -21
250 -21
251 -21
252 -21
253 -21
254 -21
255 -21
256 -21
257 -21
258 -21
259 -21
260 -21
261 -21
262 -21
263 -21
264 -21
265 -21
266 -21
267 -21
268 -21
269 -21
270 -21
271 -21
272 -21
273 -21
274 -21
275 -20
276 -21
277 -20
278 -18
279 -19
280 -21
281 -21
282 -21
283 -21
284 -21
285 -21
286 -21
287 -21
288 -21
289 -21
290 -21
291 -21
292 -21
293 -21
294 -21
295 -21
296 -21
297 -21
298 -21
299 -21
300 -21
301 -21
302 -21
303 -21
304 -21
305 -21
306 -21
307 -21
308 -21
309 -21
310 -21
311 -21
312 -21
313 -21
314 -21
315 -21
316 -21
317 -21
318 -21
319 -21
320 -21
321 -21
322 -21
323 -21
324 -21
325 -21
326 -21
327 -21
328 -21
329 -21
330 -21
331 -21
332 -21
333 -21
334 -21
335 -21
336 -21
337 -21
338 -21
339 -21
340 -21
341 -21
342 -21
343 -21
344 -21
345 -21
346 -21
347 -21
348 -21
349 -21
350 -21
351 -21
352 -21
353 -21
354 -21
355 -21
356 -21
357 -21
358 -21
359 -21
360 -21
361 -21
362 -21
363 -21
364 -21
365 -21
366 -21
367 -21
368 -21
369 -21
370 -21
371 -21
372 -21
373 -21
374 -21
375 -21
376 -21
377 -21
378 -21
379 -21
380 -21
381 -21
382 -21
383 -21
384 -21
385 -21
386 -21
387 -21
388 -21
389 -21
390 -21
391 -21
392 -21
393 -21
394 -21
395 -21
396 -21
397 -21
398 -21
399 -21
400 -21
401 -21
402 -21
403 -21
404 -21
405 -21
406 -21
407 -21
408 -21
409 -21
410 -21
411 -21
412 -21
413 -21
414 -21
415 -21
416 -21
417 -21
418 -21
419 -21
420 -21
421 -21
422 -21
423 -21
424 -21
425 -21
426 -21
427 -21
428 -21
429 -21
430 -21
431 -21
432 -21
433 -21
434 -21
435 -21
436 -21
437 -21
438 -21
439 -21
440 -21
441 -21
442 -21
443 -21
444 -21
445 -21
446 -21
447 -21
448 -21
449 -21
450 -21
451 -21
452 -21
453 -21
454 -21
455 -21
456 -21
457 -21
458 -21
459 -21
460 -21
461 -21
462 -21
463 -21
464 -21
465 -21
466 -21
467 -21
468 -21
469 -21
470 -21
471 -21
472 -21
473 -21
474 -21
475 -21
476 -21
477 -21
478 -21
479 -21
480 -21
481 -21
482 -21
483 -21
484 -21
485 -21
486 -21
487 -21
488 -21
489 -21
490 -21
491 -21
492 -21
493 -21
494 -21
495 -21
496 -21
497 -21
498 -21
499 -21
500 -21
501 -21
502 -21
503 -21
504 -21
505 -21
506 -21
507 -21
508 -21
509 -21
510 -21
511 -21
512 -21
513 -21
514 -21
515 -21
516 -21
517 -21
518 -21
519 -21
520 -21
521 -21
522 -21
523 -21
524 -21
525 -21
526 -21
527 -21
528 -21
529 -21
530 -21
531 -21
532 -21
533 -21
534 -21
535 -21
536 -21
537 -21
538 -21
539 -21
540 -21
541 -21
542 -21
543 -21
544 -21
545 -21
546 -21
547 -21
548 -21
549 -21
550 -21
551 -21
552 -21
553 -21
554 -21
555 -21
556 -21
557 -21
558 -21
559 -21
560 -21
561 -21
562 -21
563 -21
564 -21
565 -21
566 -21
567 -21
568 -21
569 -21
570 -21
571 -21
572 -21
573 -21
574 -21
575 -21
576 -21
577 -21
578 -21
579 -21
580 -21
581 -21
582 -21
583 -21
584 -21
585 -21
586 -21
587 -21
588 -21
589 -21
590 -21
591 -21
592 -21
593 -21
594 -21
595 -21
596 -21
597 -21
598 -21
599 -21
600 -21
601 -21
602 -21
603 -21
604 -21
605 -21
606 -21
607 -21
608 -21
609 -21
610 -21
611 -21
612 -21
613 -21
614 -21
615 -21
616 -21
617 -21
618 -21
619 -21
620 -21
621 -21
622 -20
623 -21
624 -21
625 -20
626 -19
627 -21
628 -21
629 -21
630 -21
631 -21
632 -21
633 -20
634 -21
635 -21
636 -19
637 -21
638 -21
639 -19
640 -21
641 -20
642 -20
643 -20
644 -21
645 -21
646 -21
647 -21
648 -19
649 -21
650 -21
651 -19
652 -21
653 -21
654 -20
655 -21
656 -21
657 -21
658 -21
659 -19
660 -21
661 -21
662 -21
663 -21
664 -21
665 -20
666 -21
667 -21
668 -20
669 -21
670 -20
671 -21
672 -21
673 -21
674 -21
675 -21
676 -21
677 -21
678 -21
679 -20
680 -20
681 -21
682 -21
683 -20
684 -21
685 -20
686 -21
687 -19
688 -21
689 -21
690 -20
691 -21
692 -21
693 -20
694 -19
695 -20
696 -21
697 -21
698 -21
699 -19
700 -21
701 -19
702 -21
703 -20
704 -21
705 -21
706 -21
707 -19
708 -21
709 -21
710 -19
711 -21
712 -20
713 -21
714 -21
715 -21
716 -20
717 -21
718 -20
719 -21
720 -21
721 -21
722 -20
723 -21
724 -20
725 -19
726 -21
727 -21
728 -18
729 -21
730 -21
731 -21
732 -20
733 -20
734 -21
735 -20
736 -21
737 -20
738 -19
739 -19
740 -19
741 -19
742 -21
743 -21
744 -21
745 -21
746 -20
747 -20
748 -21
749 -21
750 -21
751 -21
752 -20
753 -21
754 -21
755 -21
756 -20
757 -21
758 -20
759 -21
760 -21
761 -21
762 -21
763 -18
764 -21
765 -21
766 -21
767 -21
768 -21
769 -21
770 -20
771 -21
772 -20
773 -20
774 -21
775 -21
776 -21
777 -21
};
%\addlegendentry{Training}
\end{axis}

\end{tikzpicture}

%% file: fig/pong/logdir_1B_5m1_p.tex
% This file was created by tikzplotlib v0.9.2.
\begin{tikzpicture}

% \begin{axis}[
% axis background/.style={fill=white!82.7450980392157!black},
% axis line style={white},
% legend cell align={left},
% legend style={fill opacity=0.8, draw opacity=1, text opacity=1, at={(1,0.1)}, anchor=south east, draw=white!80!black},
% tick align=outside,
% tick pos=left,
% x grid style={white!69.0196078431373!black},
% xlabel={Episode},
% xmajorgrids,
% xmin=-28.4, xmax=618.4,
% xtick style={color=black},
% y grid style={white!69.0196078431373!black},
% ylabel={Reward},
% ymajorgrids,
% ymin=-22.65, ymax=13.65,
% ytick style={color=black}
% ]
\definecolor{color0}{rgb}{0.83921568627451,0.152941176470588,0.156862745098039}
\definecolor{color1}{rgb}{0.12156862745098,0.966666666666667,0.205882352941177}
\definecolor{color2}{rgb}{0.580392156862745,0.503921568627451,0.941176470588235}
\definecolor{color3}{rgb}{0.290196078431372,0.166666666666667,0.23078431372549}

\begin{axis}[
    width=\figurewidth,
    height=\figureheight,
    axis background/.style={fill=white},
    axis line style={white},
    legend cell align={left},
    legend style={fill opacity=0.8, draw opacity=1, text opacity=1, at={(1,0.1)}, anchor=south east, draw=white!80!black},
    tick align=outside,
    tick pos=left,
    x grid style={white},
    xlabel={Episode},
    ymin=-25,
    ymax=25,
    xmajorgrids,
    xtick style={color=black},
    y grid style={white!80!black},
    ylabel={Reward},
    ymajorgrids,
    ytick style={color=black},
    scaled y ticks = false,
    scaled x ticks = false
]
\addplot [semithick, color2, opacity=0.5]
table {%
1 -21
2 -21
3 -21
4 -21
5 -21
6 -19
7 -21
8 -20
9 -21
10 -21
11 -20
12 -21
13 -21
14 -21
15 -20
16 -21
17 -21
18 -21
19 -21
20 -21
21 -20
22 -19
23 -21
24 -20
25 -21
26 -21
27 -19
28 -21
29 -21
30 -20
31 -21
32 -21
33 -20
34 -21
35 -19
36 -21
37 -21
38 -20
39 -17
40 -20
41 -21
42 -21
43 -19
44 -21
45 -20
46 -21
47 -20
48 -19
49 -21
50 -20
51 -19
52 -21
53 -21
54 -21
55 -19
56 -21
57 -19
58 -17
59 -21
60 -16
61 -20
62 -17
63 -15
64 -19
65 -16
66 -13
67 -15
68 -11
69 -18
70 -17
71 -7
72 -16
73 -18
74 -15
75 -10
76 -11
77 -8
78 -10
79 -12
80 -6
81 12
82 -7
83 -5
84 -15
85 -13
86 -6
87 -6
88 3
89 -9
90 6
91 11
92 -1
93 -4
94 -9
95 -4
96 -13
97 -13
98 -9
99 -3
100 -13
101 2
102 -2
103 4
104 2
105 5
106 -6
107 -4
108 -9
109 -9
110 -3
111 -12
112 -9
113 -9
114 1
115 1
116 -10
117 -3
118 -11
119 -1
120 -9
121 -5
122 3
123 -3
124 -12
125 -18
126 -15
127 -7
128 -11
129 -5
130 -9
131 -17
132 -10
133 -16
134 -21
135 -14
136 -5
137 -16
138 -20
139 -21
140 -21
141 -20
142 -16
143 -16
144 -7
145 -9
146 -2
147 -1
148 -4
149 -19
150 -20
151 -21
152 -21
153 -21
154 -21
155 -21
156 -21
157 -21
158 -21
159 -21
160 -21
161 -21
162 -21
163 -21
164 -21
165 -21
166 -21
167 -21
168 -21
169 -21
170 -21
171 -21
172 -21
173 -21
174 -21
175 -21
176 -21
177 -21
178 -21
179 -21
180 -21
181 -21
182 -21
183 -21
184 -21
185 -21
186 -21
187 -21
188 -21
189 -21
190 -21
191 -21
192 -21
193 -21
194 -21
195 -21
196 -21
197 -21
198 -21
199 -21
200 -21
201 -21
202 -21
203 -21
204 -19
205 -21
206 -21
207 -21
208 -20
209 -19
210 -21
211 -21
212 -21
213 -21
214 -21
215 -21
216 -21
217 -21
218 -21
219 -21
220 -21
221 -21
222 -21
223 -21
224 -21
225 -21
226 -21
227 -20
228 -21
229 -21
230 -21
231 -21
232 -21
233 -21
234 -21
235 -21
236 -21
237 -21
238 -21
239 -21
240 -21
241 -21
242 -21
243 -21
244 -21
245 -21
246 -21
247 -21
248 -21
249 -21
250 -21
251 -21
252 -21
253 -21
254 -21
255 -21
256 -21
257 -21
258 -21
259 -21
260 -21
261 -21
262 -21
263 -21
264 -21
265 -21
266 -21
267 -21
268 -21
269 -21
270 -21
271 -21
272 -21
273 -21
274 -21
275 -21
276 -21
277 -21
278 -21
279 -21
280 -21
281 -21
282 -21
283 -21
284 -21
285 -21
286 -21
287 -21
288 -21
289 -21
290 -21
291 -21
292 -21
293 -21
294 -21
295 -21
296 -21
297 -21
298 -21
299 -21
300 -21
301 -21
302 -21
303 -21
304 -21
305 -21
306 -21
307 -21
308 -21
309 -21
310 -21
311 -21
312 -21
313 -21
314 -21
315 -21
316 -21
317 -21
318 -21
319 -21
320 -21
321 -21
322 -21
323 -21
324 -21
325 -21
326 -21
327 -21
328 -21
329 -21
330 -21
331 -21
332 -21
333 -21
334 -21
335 -21
336 -21
337 -21
338 -21
339 -21
340 -21
341 -21
342 -21
343 -21
344 -21
345 -21
346 -21
347 -21
348 -21
349 -21
350 -21
351 -21
352 -21
353 -21
354 -21
355 -21
356 -21
357 -21
358 -21
359 -21
360 -19
361 -21
362 -21
363 -21
364 -21
365 -21
366 -21
367 -21
368 -21
369 -21
370 -21
371 -21
372 -21
373 -21
374 -21
375 -21
376 -21
377 -21
378 -21
379 -21
380 -21
381 -21
382 -21
383 -21
384 -21
385 -21
386 -21
387 -21
388 -21
389 -21
390 -21
391 -21
392 -21
393 -21
394 -21
395 -21
396 -21
397 -21
398 -21
399 -21
400 -21
401 -21
402 -21
403 -21
404 -21
405 -21
406 -21
407 -21
408 -21
409 -21
410 -21
411 -21
412 -21
413 -21
414 -21
415 -21
416 -21
417 -21
418 -21
419 -21
420 -21
421 -21
422 -21
423 -21
424 -21
425 -21
426 -21
427 -21
428 -21
429 -21
430 -21
431 -21
432 -21
433 -21
434 -21
435 -21
436 -21
437 -21
438 -21
439 -21
440 -21
441 -21
442 -21
443 -21
444 -21
445 -21
446 -21
447 -21
448 -21
449 -21
450 -21
451 -21
452 -21
453 -21
454 -21
455 -21
456 -21
457 -21
458 -21
459 -21
460 -20
461 -21
462 -21
463 -21
464 -21
465 -21
466 -21
467 -21
468 -21
469 -21
470 -21
471 -21
472 -21
473 -21
474 -21
475 -21
476 -21
477 -21
478 -21
479 -21
480 -21
481 -21
482 -21
483 -21
484 -21
485 -21
486 -21
487 -21
488 -21
489 -21
490 -21
491 -21
492 -21
493 -21
494 -21
495 -21
496 -21
497 -21
498 -21
499 -21
500 -21
501 -21
502 -21
503 -21
504 -21
505 -21
506 -21
507 -21
508 -21
509 -21
510 -21
511 -21
512 -21
513 -21
514 -21
515 -21
516 -21
517 -21
518 -21
519 -21
520 -21
521 -21
522 -21
523 -21
524 -21
525 -21
526 -21
527 -21
528 -21
529 -21
530 -21
531 -21
532 -21
533 -21
534 -21
535 -21
536 -21
537 -21
538 -21
539 -21
540 -21
541 -21
542 -21
543 -21
544 -21
545 -21
546 -21
547 -21
548 -21
549 -21
550 -21
551 -21
552 -21
553 -21
554 -21
555 -21
556 -21
557 -21
558 -21
559 -21
560 -21
561 -19
562 -21
563 -21
564 -21
565 -21
566 -21
567 -21
568 -21
569 -21
570 -21
571 -21
572 -20
573 -21
574 -21
575 -21
576 -21
577 -21
578 -21
579 -21
580 -21
581 -21
582 -21
583 -21
584 -21
585 -21
586 -21
587 -21
588 -19
589 -20
};
%\addlegendentry{Training}
\end{axis}

\end{tikzpicture}

%% file: fig/pong/logdir_1B_10m1_p.tex
% This file was created by tikzplotlib v0.9.2.
\begin{tikzpicture}

% \begin{axis}[
% axis background/.style={fill=white!82.7450980392157!black},
% axis line style={white},
% legend cell align={left},
% legend style={fill opacity=0.8, draw opacity=1, text opacity=1, at={(1,0.1)}, anchor=south east, draw=white!80!black},
% tick align=outside,
% tick pos=left,
% x grid style={white!69.0196078431373!black},
% xlabel={Episode},
% xmajorgrids,
% xmin=-15.9, xmax=355.9,
% xtick style={color=black},
% y grid style={white!69.0196078431373!black},
% ylabel={Reward},
% ymajorgrids,
% ymin=-23, ymax=21,
% ytick style={color=black}
% ]
\definecolor{color0}{rgb}{0.83921568627451,0.152941176470588,0.156862745098039}
\definecolor{color1}{rgb}{0.12156862745098,0.966666666666667,0.205882352941177}
\definecolor{color2}{rgb}{0.580392156862745,0.503921568627451,0.941176470588235}
\definecolor{color3}{rgb}{0.290196078431372,0.166666666666667,0.23078431372549}

\begin{axis}[
    width=\figurewidth,
    height=\figureheight,
    axis background/.style={fill=white},
    axis line style={white},
    legend cell align={left},
    legend style={fill opacity=0.8, draw opacity=1, text opacity=1, at={(1,0.1)}, anchor=south east, draw=white!80!black},
    tick align=outside,
    tick pos=left,
    x grid style={white},
    xlabel={Episode},
    ymin=-25,
    ymax=25,
    xmajorgrids,
    xtick style={color=black},
    y grid style={white!80!black},
    ylabel={Reward},
    ymajorgrids,
    ytick style={color=black},
    scaled y ticks = false,
    scaled x ticks = false
]
\addplot [semithick, color2, opacity=0.5]
table {%
1 -21
2 -21
3 -21
4 -21
5 -19
6 -20
7 -21
8 -20
9 -21
10 -20
11 -21
12 -21
13 -21
14 -21
15 -21
16 -20
17 -21
18 -20
19 -21
20 -20
21 -20
22 -20
23 -20
24 -20
25 -21
26 -21
27 -20
28 -20
29 -20
30 -19
31 -20
32 -21
33 -20
34 -21
35 -21
36 -20
37 -21
38 -20
39 -21
40 -20
41 -21
42 -20
43 -21
44 -20
45 -21
46 -19
47 -20
48 -21
49 -19
50 -20
51 -21
52 -19
53 -19
54 -20
55 -21
56 -21
57 -18
58 -20
59 -20
60 -18
61 -21
62 -20
63 -20
64 -21
65 -13
66 -14
67 -17
68 -15
69 -16
70 -16
71 -9
72 -18
73 -20
74 -11
75 -11
76 -9
77 -20
78 -7
79 -6
80 -4
81 -8
82 -12
83 -11
84 -12
85 -13
86 -13
87 -6
88 3
89 -6
90 -8
91 3
92 -3
93 -8
94 -6
95 6
96 8
97 9
98 -8
99 6
100 9
101 -4
102 -3
103 7
104 4
105 7
106 14
107 14
108 14
109 17
110 18
111 17
112 18
113 15
114 17
115 16
116 16
117 13
118 17
119 15
120 15
121 16
122 13
123 9
124 18
125 12
126 17
127 13
128 13
129 14
130 8
131 15
132 15
133 17
134 12
135 14
136 17
137 13
138 -14
139 -21
140 -20
141 -21
142 -20
143 -21
144 -21
145 -21
146 -18
147 -19
148 -19
149 -21
150 -21
151 -21
152 -19
153 13
154 -14
155 14
156 -5
157 10
158 12
159 18
160 12
161 17
162 12
163 14
164 -20
165 -19
166 -20
167 -20
168 -20
169 -19
170 -21
171 -8
172 12
173 14
174 6
175 12
176 11
177 12
178 14
179 14
180 12
181 10
182 10
183 13
184 13
185 16
186 10
187 8
188 12
189 15
190 16
191 17
192 17
193 13
194 16
195 14
196 15
197 15
198 11
199 11
200 15
201 15
202 15
203 18
204 10
205 16
206 -20
207 15
208 16
209 14
210 11
211 14
212 11
213 7
214 13
215 13
216 15
217 11
218 16
219 16
220 12
221 9
222 11
223 16
224 14
225 16
226 17
227 18
228 15
229 7
230 16
231 11
232 19
233 14
234 17
235 16
236 16
237 16
238 12
239 14
240 18
241 9
242 14
243 6
244 17
245 18
246 14
247 15
248 12
249 -8
250 -7
251 10
252 14
253 14
254 15
255 15
256 17
257 14
258 15
259 16
260 12
261 13
262 15
263 15
264 12
265 15
266 17
267 13
268 18
269 15
270 18
271 17
272 14
273 15
274 15
275 13
276 16
277 18
278 15
279 4
280 15
281 13
282 -2
283 13
284 15
285 11
286 16
287 18
288 15
289 16
290 16
291 10
292 -20
293 -9
294 12
295 14
296 6
297 15
298 14
299 12
300 2
301 15
302 14
303 15
304 15
305 18
306 10
307 16
308 15
309 16
310 15
311 14
312 16
313 14
314 16
315 10
316 15
317 14
318 14
319 7
320 14
321 15
322 15
323 17
324 15
325 15
326 16
327 16
328 12
329 9
330 13
331 18
332 10
333 15
334 16
335 10
336 16
337 10
338 15
339 14
};
%\addlegendentry{Training}
\end{axis}

\end{tikzpicture}

%% file: fig/pong/logdir_1B_10m2_p.tex
% This file was created by tikzplotlib v0.9.2.
\begin{tikzpicture}

% \begin{axis}[
% axis background/.style={fill=white!82.7450980392157!black},
% axis line style={white},
% legend cell align={left},
% legend style={fill opacity=0.8, draw opacity=1, text opacity=1, at={(1,0.1)}, anchor=south east, draw=white!80!black},
% tick align=outside,
% tick pos=left,
% x grid style={white!69.0196078431373!black},
% xlabel={Episode},
% xmajorgrids,
% xmin=-12.45, xmax=283.45,
% xtick style={color=black},
% y grid style={white!69.0196078431373!black},
% ylabel={Reward},
% ymajorgrids,
% ymin=-23, ymax=21,
% ytick style={color=black}
% ]
\definecolor{color0}{rgb}{0.83921568627451,0.152941176470588,0.156862745098039}
\definecolor{color1}{rgb}{0.12156862745098,0.966666666666667,0.205882352941177}
\definecolor{color2}{rgb}{0.580392156862745,0.503921568627451,0.941176470588235}
\definecolor{color3}{rgb}{0.290196078431372,0.166666666666667,0.23078431372549}

\begin{axis}[
    width=\figurewidth,
    height=\figureheight,
    axis background/.style={fill=white},
    axis line style={white},
    legend cell align={left},
    legend style={fill opacity=0.8, draw opacity=1, text opacity=1, at={(1,0.1)}, anchor=south east, draw=white!80!black},
    tick align=outside,
    tick pos=left,
    x grid style={white},
    xlabel={Episode},
    ymin=-25,
    ymax=25,
    xmajorgrids,
    xtick style={color=black},
    y grid style={white!80!black},
    ylabel={Reward},
    ymajorgrids,
    ytick style={color=black},
    scaled y ticks = false,
    scaled x ticks = false
]
\addplot [semithick, color2, opacity=0.5]
table {%
1 -20
2 -21
3 -21
4 -21
5 -21
6 -20
7 -19
8 -18
9 -18
10 -21
11 -19
12 -19
13 -21
14 -21
15 -20
16 -20
17 -21
18 -20
19 -20
20 -20
21 -20
22 -21
23 -20
24 -20
25 -21
26 -21
27 -21
28 -21
29 -21
30 -19
31 -20
32 -21
33 -19
34 -19
35 -21
36 -21
37 -20
38 -19
39 -19
40 -19
41 -21
42 -20
43 -20
44 -21
45 -21
46 -19
47 -21
48 -20
49 -20
50 -19
51 -20
52 -21
53 -21
54 -21
55 -20
56 -19
57 -20
58 -19
59 -18
60 -19
61 -19
62 -18
63 -17
64 -20
65 -13
66 -17
67 -15
68 -16
69 -19
70 -17
71 -12
72 -21
73 -18
74 -16
75 -9
76 -12
77 -19
78 -6
79 -8
80 -8
81 -5
82 2
83 -2
84 -10
85 -10
86 -12
87 -7
88 -8
89 2
90 -3
91 -5
92 -2
93 1
94 -6
95 -6
96 -4
97 1
98 -9
99 2
100 1
101 -2
102 -6
103 -8
104 7
105 -2
106 -1
107 8
108 -10
109 4
110 -4
111 -8
112 2
113 1
114 3
115 9
116 11
117 3
118 3
119 -2
120 3
121 4
122 11
123 -3
124 7
125 -7
126 6
127 6
128 6
129 7
130 -2
131 10
132 9
133 10
134 -4
135 -3
136 9
137 8
138 9
139 8
140 7
141 -4
142 5
143 5
144 10
145 16
146 -1
147 8
148 6
149 6
150 4
151 13
152 6
153 11
154 13
155 7
156 6
157 9
158 13
159 6
160 14
161 14
162 16
163 13
164 14
165 7
166 3
167 6
168 8
169 11
170 10
171 3
172 11
173 12
174 13
175 13
176 13
177 4
178 7
179 12
180 13
181 11
182 12
183 7
184 12
185 14
186 14
187 14
188 1
189 6
190 7
191 13
192 11
193 10
194 6
195 13
196 13
197 9
198 5
199 12
200 14
201 10
202 12
203 15
204 9
205 10
206 6
207 2
208 6
209 11
210 15
211 17
212 17
213 17
214 18
215 16
216 18
217 16
218 18
219 17
220 14
221 18
222 18
223 17
224 17
225 12
226 15
227 16
228 16
229 19
230 14
231 17
232 17
233 18
234 17
235 14
236 16
237 18
238 16
239 12
240 13
241 15
242 16
243 17
244 16
245 14
246 16
247 17
248 15
249 15
250 18
251 17
252 3
253 14
254 16
255 16
256 14
257 17
258 18
259 15
260 17
261 18
262 16
263 8
264 16
265 14
266 16
267 18
268 15
269 16
270 17
};
%\addlegendentry{Training}
\end{axis}

\end{tikzpicture}

%% file: fig/pong/logdir_2B_10m2_p.tex
% This file was created by tikzplotlib v0.9.2.
\begin{tikzpicture}

% \begin{axis}[
% axis background/.style={fill=white!82.7450980392157!black},
% axis line style={white},
% legend cell align={left},
% legend style={fill opacity=0.8, draw opacity=1, text opacity=1, at={(1,0.1)}, anchor=south east, draw=white!80!black},
% tick align=outside,
% tick pos=left,
% x grid style={white!69.0196078431373!black},
% xlabel={Episode},
% xmajorgrids,
% xmin=-12.85, xmax=291.85,
% xtick style={color=black},
% y grid style={white!69.0196078431373!black},
% ylabel={Reward},
% ymajorgrids,
% ymin=-22.95, ymax=19.95,
% ytick style={color=black}
% ]
\definecolor{color0}{rgb}{0.83921568627451,0.152941176470588,0.156862745098039}
\definecolor{color1}{rgb}{0.12156862745098,0.966666666666667,0.205882352941177}
\definecolor{color2}{rgb}{0.580392156862745,0.503921568627451,0.941176470588235}
\definecolor{color3}{rgb}{0.290196078431372,0.166666666666667,0.23078431372549}

\begin{axis}[
    width=\figurewidth,
    height=\figureheight,
    axis background/.style={fill=white},
    axis line style={white},
    legend cell align={left},
    legend style={fill opacity=0.8, draw opacity=1, text opacity=1, at={(1,0.1)}, anchor=south east, draw=white!80!black},
    tick align=outside,
    tick pos=left,
    x grid style={white},
    xlabel={Episode},
    ymin=-25,
    ymax=25,
    xmajorgrids,
    xtick style={color=black},
    y grid style={white!80!black},
    ylabel={Reward},
    ymajorgrids,
    ytick style={color=black},
    scaled y ticks = false,
    scaled x ticks = false
]
\addplot [semithick, color2, opacity=0.5]
table {%
1 -17
2 -21
3 -19
4 -21
5 -20
6 -19
7 -21
8 -21
9 -20
10 -20
11 -20
12 -21
13 -21
14 -21
15 -20
16 -21
17 -21
18 -20
19 -21
20 -21
21 -20
22 -20
23 -20
24 -21
25 -21
26 -21
27 -20
28 -21
29 -21
30 -19
31 -19
32 -20
33 -21
34 -20
35 -20
36 -20
37 -20
38 -19
39 -20
40 -21
41 -20
42 -19
43 -20
44 -18
45 -21
46 -19
47 -20
48 -20
49 -20
50 -19
51 -19
52 -21
53 -19
54 -19
55 -19
56 -21
57 -16
58 -21
59 -20
60 -18
61 -19
62 -19
63 -19
64 -18
65 -16
66 -15
67 -13
68 -12
69 -19
70 -19
71 -6
72 -18
73 -14
74 -8
75 -11
76 -9
77 -18
78 -3
79 -17
80 -20
81 -12
82 -4
83 -5
84 -6
85 -16
86 -8
87 -6
88 -8
89 12
90 2
91 -1
92 3
93 6
94 -7
95 -5
96 -20
97 4
98 -2
99 4
100 4
101 -5
102 -21
103 -21
104 -21
105 -9
106 12
107 6
108 1
109 2
110 8
111 5
112 4
113 2
114 1
115 9
116 12
117 1
118 13
119 1
120 13
121 -6
122 12
123 13
124 -8
125 13
126 11
127 -3
128 12
129 -2
130 14
131 9
132 2
133 11
134 -1
135 9
136 16
137 4
138 -2
139 -3
140 13
141 12
142 -8
143 -3
144 -1
145 9
146 10
147 -5
148 8
149 7
150 9
151 -3
152 9
153 5
154 8
155 18
156 7
157 15
158 16
159 1
160 9
161 4
162 2
163 10
164 -5
165 5
166 -10
167 -8
168 -8
169 2
170 4
171 14
172 10
173 3
174 14
175 11
176 -1
177 13
178 10
179 5
180 9
181 -1
182 12
183 10
184 14
185 7
186 -4
187 12
188 12
189 12
190 7
191 17
192 9
193 6
194 14
195 2
196 -14
197 -2
198 5
199 9
200 9
201 10
202 8
203 8
204 12
205 7
206 -1
207 6
208 2
209 11
210 11
211 3
212 -1
213 8
214 14
215 12
216 11
217 14
218 12
219 14
220 13
221 7
222 5
223 1
224 4
225 -2
226 8
227 9
228 9
229 12
230 11
231 10
232 6
233 6
234 14
235 15
236 15
237 13
238 5
239 3
240 -14
241 -21
242 -20
243 3
244 3
245 9
246 10
247 16
248 14
249 -3
250 5
251 9
252 11
253 11
254 7
255 7
256 1
257 -7
258 -6
259 17
260 10
261 13
262 -6
263 -8
264 16
265 7
266 13
267 8
268 10
269 13
270 12
271 12
272 14
273 8
274 2
275 5
276 9
277 -7
278 10
};
%\addlegendentry{Training}
\end{axis}

\end{tikzpicture}

%% file: fig/pong/logdir_4B_20m2_p.tex
% This file was created by tikzplotlib v0.9.2.
\begin{tikzpicture}

% \begin{axis}[
% axis background/.style={fill=white!82.7450980392157!black},
% axis line style={white},
% legend cell align={left},
% legend style={fill opacity=0.8, draw opacity=1, text opacity=1, at={(1,0.1)}, anchor=south east, draw=white!80!black},
% tick align=outside,
% tick pos=left,
% x grid style={white!69.0196078431373!black},
% xlabel={Episode},
% xmajorgrids,
% xmin=-15.6, xmax=349.6,
% xtick style={color=black},
% y grid style={white!69.0196078431373!black},
% ylabel={Reward},
% ymajorgrids,
% ymin=-23.1, ymax=23.1,
% ytick style={color=black}
% ]
\definecolor{color0}{rgb}{0.83921568627451,0.152941176470588,0.156862745098039}
\definecolor{color1}{rgb}{0.12156862745098,0.966666666666667,0.205882352941177}
\definecolor{color2}{rgb}{0.580392156862745,0.503921568627451,0.941176470588235}
\definecolor{color3}{rgb}{0.290196078431372,0.166666666666667,0.23078431372549}

\begin{axis}[
    width=\figurewidth,
    height=\figureheight,
    axis background/.style={fill=white},
    axis line style={white},
    legend cell align={left},
    legend style={fill opacity=0.8, draw opacity=1, text opacity=1, at={(1,0.1)}, anchor=south east, draw=white!80!black},
    tick align=outside,
    tick pos=left,
    x grid style={white},
    xlabel={Episode},
    ymin=-25,
    ymax=25,
    xmajorgrids,
    xtick style={color=black},
    y grid style={white!80!black},
    ylabel={Reward},
    ymajorgrids,
    ytick style={color=black},
    scaled y ticks = false,
    scaled x ticks = false
]
\addplot [semithick, color2, opacity=0.5]
table {%
1 -21
2 -21
3 -19
4 -21
5 -19
6 -20
7 -20
8 -20
9 -19
10 -20
11 -20
12 -18
13 -20
14 -21
15 -18
16 -21
17 -21
18 -21
19 -20
20 -21
21 -21
22 -20
23 -21
24 -21
25 -20
26 -21
27 -20
28 -21
29 -20
30 -21
31 -21
32 -20
33 -21
34 -21
35 -21
36 -20
37 -21
38 -18
39 -20
40 -21
41 -21
42 -19
43 -20
44 -19
45 -21
46 -21
47 -20
48 -19
49 -20
50 -21
51 -20
52 -20
53 -19
54 -21
55 -19
56 -19
57 -18
58 -20
59 -19
60 -16
61 -18
62 -20
63 -20
64 -17
65 -17
66 -13
67 -18
68 -16
69 -21
70 -13
71 -12
72 -20
73 -21
74 -15
75 -8
76 -9
77 -11
78 -9
79 -4
80 -7
81 -8
82 -7
83 -9
84 -5
85 -13
86 -7
87 -2
88 -3
89 5
90 1
91 -2
92 -6
93 -8
94 -5
95 -3
96 3
97 -17
98 -21
99 -9
100 1
101 -3
102 -1
103 -5
104 -1
105 -2
106 -1
107 9
108 -5
109 15
110 5
111 -5
112 9
113 13
114 19
115 11
116 7
117 13
118 17
119 15
120 15
121 8
122 16
123 16
124 15
125 18
126 16
127 16
128 14
129 18
130 13
131 15
132 20
133 -18
134 12
135 16
136 16
137 19
138 13
139 9
140 14
141 17
142 17
143 18
144 17
145 15
146 13
147 10
148 12
149 13
150 19
151 6
152 15
153 17
154 -2
155 20
156 14
157 19
158 15
159 16
160 14
161 16
162 20
163 15
164 10
165 16
166 13
167 17
168 13
169 17
170 16
171 12
172 17
173 15
174 17
175 17
176 16
177 13
178 14
179 17
180 15
181 14
182 14
183 17
184 19
185 16
186 18
187 15
188 11
189 10
190 18
191 9
192 13
193 18
194 14
195 18
196 5
197 18
198 11
199 16
200 14
201 17
202 17
203 15
204 19
205 12
206 19
207 7
208 8
209 5
210 13
211 -13
212 18
213 8
214 18
215 16
216 15
217 10
218 -17
219 8
220 10
221 -7
222 12
223 14
224 17
225 14
226 13
227 14
228 16
229 15
230 17
231 14
232 12
233 17
234 18
235 15
236 18
237 17
238 9
239 12
240 14
241 5
242 4
243 3
244 10
245 14
246 18
247 14
248 16
249 17
250 15
251 -4
252 17
253 11
254 18
255 18
256 15
257 15
258 19
259 17
260 16
261 10
262 15
263 13
264 15
265 16
266 20
267 19
268 18
269 8
270 14
271 16
272 18
273 18
274 17
275 -14
276 7
277 17
278 15
279 16
280 19
281 8
282 14
283 17
284 17
285 17
286 14
287 15
288 12
289 -21
290 15
291 17
292 14
293 16
294 15
295 8
296 12
297 7
298 1
299 12
300 14
301 -4
302 -21
303 -19
304 -17
305 18
306 -20
307 17
308 18
309 21
310 -6
311 -20
312 16
313 12
314 14
315 18
316 18
317 16
318 17
319 15
320 14
321 16
322 16
323 17
324 15
325 19
326 16
327 14
328 11
329 16
330 9
331 5
332 15
333 8
};
%\addlegendentry{Training}
\end{axis}

\end{tikzpicture}

%% file: fig/breakout/logdir_0B_0713.tex
% This file was created by tikzplotlib v0.9.2.
\begin{tikzpicture}

% \begin{axis}[
% axis background/.style={fill=white!82.7450980392157!black},
% axis line style={white},
% legend cell align={left},
% legend style={fill opacity=0.8, draw opacity=1, text opacity=1, at={(1,0.1)}, anchor=south east, draw=white!80!black},
% tick align=outside,
% tick pos=left,
% x grid style={white!69.0196078431373!black},
% xlabel={Episode},
% xmajorgrids,
% xmin=-20.75, xmax=457.75,
% xtick style={color=black},
% y grid style={white!69.0196078431373!black},
% ylabel={Reward},
% ymajorgrids,
% ymin=-42.65, ymax=895.65,
% ytick style={color=black}
% ]
\definecolor{color0}{rgb}{0.83921568627451,0.152941176470588,0.156862745098039}
\definecolor{color1}{rgb}{0.12156862745098,0.966666666666667,0.205882352941177}
\definecolor{color2}{rgb}{0.580392156862745,0.503921568627451,0.941176470588235}
\definecolor{color3}{rgb}{0.290196078431372,0.166666666666667,0.23078431372549}

\begin{axis}[
    width=\figurewidth,
    height=\figureheight,
    axis background/.style={fill=white},
    axis line style={white},
    legend cell align={left},
    legend style={fill opacity=0.8, draw opacity=1, text opacity=1, at={(1,0.1)}, anchor=south east, draw=white!80!black},
    tick align=outside,
    tick pos=left,
    x grid style={white},
    xlabel={Episode},
    ymin=-20,
    ymax=420,
    xmajorgrids,
    xtick style={color=black},
    y grid style={white!80!black},
    ylabel={Reward},
    ymajorgrids,
    ytick style={color=black},
    scaled y ticks = false,
    scaled x ticks = false
]
\addplot [semithick, color3, opacity=0.3]
table {%
1 0
2 4
3 0
4 1
5 1
6 3
7 0
8 4
9 2
10 2
11 0
12 0
13 4
14 5
15 5
16 0
17 0
18 0
19 0
20 2
21 0
22 9
23 0
24 0
25 3
26 0
27 0
28 0
29 0
30 5
31 10
32 5
33 2
34 0
35 2
36 5
37 0
38 0
39 4
40 2
41 5
42 3
43 1
44 0
45 5
46 0
47 4
48 1
49 2
50 2
51 7
52 2
53 1
54 3
55 0
56 2
57 2
58 0
59 8
60 10
61 2
62 4
63 3
64 3
65 3
66 2
67 15
68 1
69 1
70 0
71 6
72 0
73 12
74 8
75 4
76 4
77 6
78 7
79 7
80 2
81 8
82 5
83 8
84 7
85 12
86 8
87 9
88 10
89 17
90 10
91 12
92 10
93 12
94 16
95 17
96 16
97 13
98 17
99 14
100 21
101 13
102 31
103 34
104 22
105 23
106 42
107 14
108 33
109 26
110 9
111 23
112 18
113 30
114 25
115 54
116 27
117 31
118 27
119 36
120 43
121 19
122 26
123 73
124 20
125 26
126 71
127 73
128 84
129 83
130 37
131 50
132 33
133 15
134 48
135 102
136 32
137 63
138 45
139 55
140 28
141 135
142 67
143 103
144 103
145 269
146 34
147 80
148 124
149 69
150 52
151 85
152 55
153 68
154 32
155 62
156 63
157 83
158 52
159 94
160 217
161 92
162 129
163 32
164 100
165 93
166 141
167 79
168 108
169 66
170 120
171 143
172 57
173 54
174 74
175 26
176 114
177 227
178 118
179 336
180 69
181 80
182 226
183 150
184 119
185 21
186 285
187 365
188 54
189 117
190 63
191 406
192 114
193 97
194 92
195 203
196 267
197 236
198 74
199 288
200 64
201 115
202 125
203 51
204 37
205 217
206 43
207 118
208 235
209 221
210 424
211 384
212 194
213 359
214 221
215 244
216 145
217 250
218 157
219 351
220 377
221 215
222 355
223 88
224 384
225 151
226 254
227 194
228 357
229 380
230 297
231 183
232 378
233 355
234 277
235 385
236 380
237 383
238 358
239 338
240 58
241 81
242 363
243 358
244 58
245 121
246 371
247 180
248 378
249 225
250 298
251 72
252 158
253 239
254 377
255 379
256 370
257 412
258 354
259 351
260 310
261 390
262 421
263 155
264 322
265 404
266 423
267 231
268 368
269 74
270 337
271 95
272 357
273 210
274 413
275 404
276 321
277 108
278 391
279 294
280 354
281 94
282 385
283 375
284 245
285 39
286 280
287 352
288 387
289 410
290 262
291 468
292 367
293 367
294 364
295 368
296 347
297 359
298 372
299 390
300 134
301 7
302 338
303 384
304 413
305 327
306 390
307 349
308 90
309 412
310 352
311 421
312 432
313 386
314 70
315 0
316 367
317 385
318 344
319 64
320 343
321 475
322 409
323 327
324 399
325 394
326 404
327 367
328 401
329 409
330 363
331 402
332 405
333 384
334 379
335 334
336 381
337 428
338 411
339 241
340 276
341 359
342 325
343 417
344 416
345 409
346 286
347 413
348 373
349 406
350 409
351 426
352 314
353 90
354 363
355 418
356 406
357 364
358 368
359 180
360 424
361 389
362 112
363 348
364 340
365 385
366 405
367 411
368 423
369 37
370 377
371 345
372 403
373 406
374 355
375 281
376 404
377 333
378 417
379 336
380 75
381 2
382 0
383 235
384 415
385 374
386 296
387 218
388 336
389 120
390 375
391 76
392 853
393 427
394 306
395 287
396 388
397 415
398 470
399 424
400 405
401 415
402 56
403 344
404 428
405 410
406 425
407 386
408 394
409 415
410 420
411 427
412 406
413 405
414 382
415 394
416 359
417 388
418 427
419 335
420 420
421 333
422 181
423 44
424 424
425 405
426 402
427 378
428 391
429 370
430 281
431 352
432 397
433 420
434 205
435 109
436 413
};
%\addlegendentry{Training}
\end{axis}

\end{tikzpicture}

%% file: fig/breakout/logdir_1B_10m1_p.tex
% This file was created by tikzplotlib v0.9.2.
\begin{tikzpicture}

% \begin{axis}[
% axis background/.style={fill=white!82.7450980392157!black},
% axis line style={white},
% legend cell align={left},
% legend style={fill opacity=0.8, draw opacity=1, text opacity=1, at={(1,0.1)}, anchor=south east, draw=white!80!black},
% tick align=outside,
% tick pos=left,
% x grid style={white!69.0196078431373!black},
% xlabel={Episode},
% xmajorgrids,
% xmin=-43.45, xmax=934.45,
% xtick style={color=black},
% y grid style={white!69.0196078431373!black},
% ylabel={Reward},
% ymajorgrids,
% ymin=-17.95, ymax=376.95,
% ytick style={color=black}
% ]
\definecolor{color0}{rgb}{0.83921568627451,0.152941176470588,0.156862745098039}
\definecolor{color1}{rgb}{0.12156862745098,0.966666666666667,0.205882352941177}
\definecolor{color2}{rgb}{0.580392156862745,0.503921568627451,0.941176470588235}
\definecolor{color3}{rgb}{0.290196078431372,0.166666666666667,0.23078431372549}

\begin{axis}[
    width=\figurewidth,
    height=\figureheight,
    axis background/.style={fill=white},
    axis line style={white},
    legend cell align={left},
    legend style={fill opacity=0.8, draw opacity=1, text opacity=1, at={(1,0.1)}, anchor=south east, draw=white!80!black},
    tick align=outside,
    tick pos=left,
    x grid style={white},
    xlabel={Episode},
    ymin=-20,
    ymax=420,
    xmajorgrids,
    xtick style={color=black},
    y grid style={white!80!black},
    ylabel={Reward},
    ymajorgrids,
    ytick style={color=black},
    scaled y ticks = false,
    scaled x ticks = false
]
\addplot [semithick, color3, opacity=0.3]
table {%
1 0
2 2
3 2
4 1
5 2
6 1
7 1
8 0
9 0
10 2
11 1
12 4
13 2
14 2
15 11
16 0
17 2
18 0
19 0
20 3
21 0
22 4
23 1
24 0
25 2
26 0
27 0
28 3
29 0
30 6
31 11
32 9
33 6
34 2
35 0
36 9
37 0
38 1
39 0
40 0
41 9
42 6
43 0
44 1
45 1
46 1
47 5
48 0
49 0
50 1
51 2
52 0
53 0
54 5
55 1
56 2
57 0
58 0
59 0
60 11
61 1
62 7
63 1
64 8
65 0
66 1
67 16
68 6
69 6
70 6
71 1
72 2
73 4
74 8
75 5
76 6
77 6
78 14
79 13
80 9
81 13
82 10
83 6
84 9
85 14
86 5
87 22
88 16
89 2
90 6
91 16
92 8
93 12
94 10
95 19
96 14
97 4
98 14
99 17
100 8
101 8
102 14
103 19
104 16
105 7
106 20
107 34
108 19
109 21
110 9
111 22
112 8
113 13
114 9
115 23
116 18
117 12
118 35
119 26
120 9
121 36
122 38
123 8
124 25
125 23
126 23
127 26
128 22
129 13
130 37
131 31
132 28
133 39
134 25
135 45
136 12
137 41
138 17
139 27
140 17
141 27
142 29
143 27
144 23
145 32
146 40
147 12
148 26
149 31
150 28
151 70
152 25
153 19
154 58
155 19
156 26
157 16
158 30
159 31
160 4
161 48
162 57
163 22
164 45
165 26
166 62
167 75
168 24
169 0
170 22
171 23
172 24
173 42
174 17
175 76
176 63
177 8
178 40
179 60
180 33
181 51
182 22
183 83
184 5
185 0
186 325
187 18
188 3
189 50
190 7
191 17
192 55
193 16
194 37
195 85
196 52
197 64
198 42
199 67
200 14
201 15
202 50
203 70
204 75
205 59
206 82
207 48
208 7
209 24
210 38
211 76
212 61
213 32
214 35
215 37
216 54
217 49
218 17
219 61
220 48
221 43
222 10
223 60
224 52
225 96
226 103
227 21
228 52
229 21
230 31
231 42
232 67
233 66
234 16
235 43
236 42
237 38
238 32
239 92
240 50
241 52
242 25
243 23
244 68
245 21
246 68
247 58
248 101
249 10
250 67
251 31
252 47
253 68
254 27
255 51
256 85
257 25
258 34
259 46
260 39
261 31
262 26
263 28
264 45
265 28
266 31
267 23
268 20
269 16
270 49
271 8
272 46
273 84
274 44
275 18
276 78
277 39
278 15
279 51
280 22
281 69
282 11
283 149
284 45
285 45
286 21
287 36
288 41
289 36
290 28
291 56
292 28
293 54
294 53
295 5
296 57
297 16
298 42
299 16
300 24
301 57
302 29
303 71
304 77
305 37
306 33
307 51
308 56
309 13
310 22
311 46
312 28
313 43
314 43
315 1
316 0
317 2
318 0
319 17
320 68
321 13
322 41
323 37
324 31
325 13
326 25
327 70
328 5
329 23
330 21
331 12
332 11
333 19
334 11
335 16
336 37
337 15
338 20
339 27
340 46
341 14
342 18
343 29
344 57
345 20
346 40
347 50
348 5
349 30
350 55
351 8
352 43
353 34
354 71
355 17
356 13
357 16
358 31
359 56
360 7
361 48
362 17
363 42
364 65
365 91
366 52
367 38
368 92
369 19
370 29
371 113
372 28
373 47
374 71
375 78
376 80
377 46
378 62
379 20
380 56
381 27
382 30
383 20
384 26
385 13
386 17
387 30
388 72
389 93
390 21
391 64
392 62
393 22
394 37
395 58
396 41
397 18
398 44
399 27
400 75
401 69
402 136
403 116
404 48
405 16
406 12
407 46
408 20
409 39
410 33
411 101
412 62
413 72
414 42
415 33
416 359
417 160
418 15
419 26
420 22
421 44
422 39
423 21
424 257
425 91
426 49
427 48
428 27
429 221
430 111
431 7
432 19
433 62
434 21
435 21
436 23
437 68
438 37
439 30
440 57
441 10
442 306
443 34
444 24
445 18
446 50
447 38
448 99
449 20
450 32
451 17
452 52
453 33
454 98
455 54
456 68
457 38
458 86
459 17
460 38
461 49
462 58
463 18
464 102
465 93
466 38
467 29
468 47
469 30
470 35
471 322
472 28
473 30
474 52
475 26
476 200
477 35
478 95
479 39
480 18
481 3
482 16
483 41
484 112
485 186
486 47
487 19
488 5
489 0
490 9
491 14
492 16
493 68
494 46
495 29
496 44
497 16
498 35
499 63
500 46
501 41
502 22
503 29
504 30
505 19
506 170
507 153
508 141
509 27
510 18
511 158
512 27
513 41
514 66
515 36
516 44
517 71
518 23
519 18
520 221
521 71
522 59
523 28
524 46
525 48
526 10
527 35
528 18
529 32
530 34
531 50
532 27
533 88
534 28
535 13
536 41
537 57
538 51
539 22
540 10
541 11
542 22
543 64
544 37
545 24
546 104
547 31
548 58
549 29
550 34
551 78
552 47
553 4
554 78
555 43
556 58
557 0
558 33
559 132
560 46
561 11
562 10
563 2
564 6
565 14
566 17
567 0
568 6
569 66
570 33
571 57
572 17
573 32
574 53
575 40
576 5
577 11
578 0
579 0
580 4
581 0
582 2
583 0
584 2
585 9
586 0
587 13
588 46
589 25
590 6
591 22
592 11
593 37
594 18
595 31
596 22
597 59
598 21
599 29
600 42
601 20
602 9
603 70
604 26
605 56
606 42
607 10
608 101
609 47
610 8
611 39
612 22
613 38
614 53
615 74
616 67
617 49
618 3
619 41
620 49
621 65
622 58
623 14
624 25
625 26
626 47
627 65
628 17
629 63
630 13
631 26
632 9
633 55
634 54
635 39
636 77
637 19
638 53
639 63
640 82
641 9
642 49
643 55
644 38
645 29
646 48
647 4
648 21
649 48
650 28
651 17
652 23
653 64
654 49
655 43
656 62
657 48
658 18
659 67
660 35
661 49
662 37
663 14
664 49
665 7
666 39
667 23
668 0
669 34
670 22
671 62
672 18
673 17
674 12
675 74
676 17
677 18
678 40
679 114
680 33
681 40
682 51
683 9
684 29
685 39
686 43
687 28
688 36
689 37
690 17
691 8
692 29
693 22
694 33
695 52
696 12
697 12
698 22
699 18
700 32
701 9
702 4
703 11
704 0
705 0
706 11
707 0
708 0
709 0
710 0
711 11
712 0
713 0
714 0
715 0
716 0
717 11
718 0
719 0
720 9
721 2
722 0
723 0
724 0
725 0
726 0
727 2
728 0
729 2
730 0
731 0
732 7
733 11
734 0
735 0
736 11
737 4
738 11
739 11
740 22
741 5
742 22
743 40
744 20
745 29
746 64
747 7
748 18
749 21
750 5
751 9
752 23
753 0
754 1
755 0
756 0
757 0
758 1
759 1
760 0
761 0
762 2
763 3
764 0
765 1
766 2
767 0
768 0
769 0
770 2
771 2
772 5
773 4
774 6
775 22
776 15
777 60
778 28
779 89
780 13
781 29
782 63
783 50
784 30
785 48
786 30
787 31
788 49
789 9
790 28
791 62
792 8
793 1
794 1
795 2
796 1
797 0
798 0
799 1
800 0
801 3
802 0
803 7
804 0
805 7
806 19
807 36
808 22
809 16
810 21
811 31
812 21
813 19
814 16
815 30
816 43
817 6
818 9
819 0
820 0
821 10
822 0
823 0
824 0
825 0
826 0
827 0
828 11
829 0
830 0
831 0
832 0
833 0
834 11
835 0
836 0
837 0
838 0
839 1
840 0
841 0
842 11
843 1
844 11
845 0
846 0
847 0
848 0
849 0
850 0
851 0
852 0
853 0
854 0
855 3
856 3
857 0
858 11
859 0
860 0
861 7
862 1
863 0
864 0
865 0
866 9
867 9
868 7
869 12
870 4
871 17
872 11
873 30
874 19
875 11
876 27
877 36
878 15
879 17
880 21
881 37
882 24
883 36
884 17
885 56
886 69
887 10
888 59
889 48
890 45
};
%\addlegendentry{Training}
\end{axis}

\end{tikzpicture}

%% file: fig/breakout/logdir_1B_10m2_p.tex
% This file was created by tikzplotlib v0.9.2.
\begin{tikzpicture}

% \begin{axis}[
% axis background/.style={fill=white!82.7450980392157!black},
% axis line style={white},
% legend cell align={left},
% legend style={fill opacity=0.8, draw opacity=1, text opacity=1, at={(1,0.1)}, anchor=south east, draw=white!80!black},
% tick align=outside,
% tick pos=left,
% x grid style={white!69.0196078431373!black},
% xlabel={Episode},
% xmajorgrids,
% xmin=-27.95, xmax=608.95,
% xtick style={color=black},
% y grid style={white!69.0196078431373!black},
% ylabel={Reward},
% ymajorgrids,
% ymin=-21.3, ymax=447.3,
% ytick style={color=black}
% ]
\definecolor{color0}{rgb}{0.83921568627451,0.152941176470588,0.156862745098039}
\definecolor{color1}{rgb}{0.12156862745098,0.966666666666667,0.205882352941177}
\definecolor{color2}{rgb}{0.580392156862745,0.503921568627451,0.941176470588235}
\definecolor{color3}{rgb}{0.290196078431372,0.166666666666667,0.23078431372549}

\begin{axis}[
    width=\figurewidth,
    height=\figureheight,
    axis background/.style={fill=white},
    axis line style={white},
    legend cell align={left},
    legend style={fill opacity=0.8, draw opacity=1, text opacity=1, at={(1,0.1)}, anchor=south east, draw=white!80!black},
    tick align=outside,
    tick pos=left,
    x grid style={white},
    xlabel={Episode},
    ymin=-20,
    ymax=420,
    xmajorgrids,
    xtick style={color=black},
    y grid style={white!80!black},
    ylabel={Reward},
    ymajorgrids,
    ytick style={color=black},
    scaled y ticks = false,
    scaled x ticks = false
]
\addplot [semithick, color3, opacity=0.3]
table {%
1 0
2 4
3 2
4 2
5 0
6 11
7 0
8 2
9 1
10 2
11 4
12 2
13 4
14 3
15 5
16 0
17 3
18 0
19 1
20 0
21 0
22 5
23 0
24 0
25 2
26 0
27 0
28 0
29 0
30 4
31 9
32 6
33 2
34 2
35 2
36 9
37 1
38 0
39 3
40 1
41 7
42 12
43 3
44 0
45 4
46 5
47 6
48 2
49 1
50 2
51 0
52 1
53 2
54 0
55 0
56 3
57 0
58 2
59 0
60 4
61 1
62 7
63 1
64 2
65 1
66 1
67 4
68 2
69 1
70 4
71 0
72 2
73 4
74 1
75 0
76 0
77 5
78 10
79 9
80 4
81 7
82 7
83 7
84 11
85 6
86 11
87 5
88 9
89 9
90 20
91 5
92 10
93 7
94 14
95 8
96 18
97 9
98 10
99 9
100 8
101 15
102 19
103 24
104 36
105 11
106 8
107 16
108 11
109 17
110 16
111 25
112 11
113 22
114 14
115 18
116 29
117 14
118 33
119 30
120 51
121 75
122 28
123 22
124 27
125 42
126 57
127 35
128 11
129 20
130 15
131 15
132 30
133 40
134 47
135 48
136 33
137 52
138 39
139 80
140 47
141 96
142 45
143 42
144 61
145 78
146 51
147 71
148 58
149 31
150 63
151 39
152 108
153 30
154 37
155 30
156 71
157 73
158 44
159 45
160 53
161 45
162 230
163 24
164 63
165 43
166 97
167 39
168 84
169 51
170 88
171 27
172 52
173 61
174 49
175 294
176 39
177 98
178 33
179 111
180 79
181 18
182 104
183 81
184 53
185 109
186 140
187 149
188 63
189 52
190 46
191 52
192 62
193 89
194 56
195 152
196 32
197 103
198 266
199 190
200 60
201 48
202 127
203 107
204 38
205 14
206 61
207 36
208 108
209 117
210 58
211 45
212 69
213 150
214 106
215 111
216 332
217 75
218 52
219 95
220 20
221 78
222 69
223 19
224 115
225 181
226 124
227 266
228 74
229 88
230 184
231 41
232 99
233 43
234 30
235 70
236 337
237 40
238 30
239 261
240 169
241 217
242 41
243 26
244 97
245 198
246 37
247 120
248 84
249 48
250 0
251 42
252 101
253 189
254 31
255 22
256 110
257 117
258 76
259 298
260 29
261 270
262 118
263 57
264 307
265 383
266 76
267 49
268 26
269 122
270 42
271 67
272 22
273 364
274 104
275 126
276 40
277 371
278 194
279 48
280 61
281 314
282 71
283 90
284 366
285 143
286 166
287 87
288 54
289 95
290 188
291 126
292 46
293 39
294 89
295 87
296 130
297 109
298 53
299 63
300 133
301 79
302 12
303 74
304 52
305 16
306 0
307 9
308 34
309 312
310 187
311 98
312 83
313 34
314 109
315 401
316 72
317 71
318 133
319 50
320 35
321 78
322 23
323 55
324 127
325 305
326 69
327 237
328 43
329 68
330 276
331 77
332 108
333 60
334 100
335 85
336 5
337 0
338 75
339 54
340 21
341 290
342 27
343 60
344 152
345 121
346 93
347 247
348 94
349 63
350 85
351 137
352 202
353 17
354 3
355 3
356 61
357 35
358 132
359 104
360 335
361 213
362 83
363 184
364 43
365 49
366 118
367 79
368 351
369 399
370 78
371 65
372 167
373 402
374 329
375 49
376 267
377 14
378 126
379 55
380 118
381 166
382 52
383 78
384 62
385 66
386 37
387 72
388 100
389 421
390 59
391 105
392 66
393 153
394 88
395 179
396 150
397 55
398 56
399 258
400 412
401 108
402 38
403 378
404 48
405 179
406 85
407 41
408 102
409 22
410 47
411 140
412 268
413 238
414 51
415 55
416 49
417 50
418 369
419 189
420 53
421 75
422 100
423 426
424 51
425 76
426 72
427 329
428 102
429 94
430 147
431 104
432 61
433 370
434 29
435 311
436 108
437 70
438 86
439 43
440 323
441 181
442 62
443 66
444 80
445 335
446 117
447 32
448 239
449 165
450 54
451 407
452 127
453 73
454 29
455 366
456 63
457 103
458 46
459 68
460 26
461 37
462 159
463 210
464 36
465 69
466 97
467 25
468 30
469 218
470 52
471 227
472 142
473 115
474 354
475 49
476 15
477 25
478 393
479 87
480 350
481 132
482 41
483 301
484 38
485 73
486 254
487 85
488 89
489 165
490 342
491 88
492 87
493 51
494 52
495 38
496 390
497 14
498 138
499 87
500 63
501 129
502 77
503 24
504 76
505 65
506 61
507 49
508 42
509 155
510 48
511 50
512 74
513 214
514 42
515 86
516 79
517 172
518 23
519 18
520 421
521 49
522 280
523 30
524 395
525 79
526 267
527 90
528 135
529 245
530 107
531 54
532 36
533 47
534 45
535 66
536 157
537 16
538 362
539 90
540 408
541 52
542 243
543 391
544 162
545 37
546 64
547 380
548 62
549 102
550 103
551 177
552 115
553 166
554 203
555 259
556 79
557 352
558 382
559 50
560 47
561 387
562 75
563 60
564 93
565 214
566 55
567 262
568 57
569 375
570 302
571 17
572 96
573 166
574 86
575 76
576 313
577 67
578 67
579 408
580 392
};
%\addlegendentry{Training}
\end{axis}

\end{tikzpicture}

%% file: fig/breakout/logdir_1B_20m2_p.tex
% This file was created by tikzplotlib v0.9.2.
\begin{tikzpicture}

% \begin{axis}[
% axis background/.style={fill=white!82.7450980392157!black},
% axis line style={white},
% legend cell align={left},
% legend style={fill opacity=0.8, draw opacity=1, text opacity=1, at={(1,0.1)}, anchor=south east, draw=white!80!black},
% tick align=outside,
% tick pos=left,
% x grid style={white!69.0196078431373!black},
% xlabel={Episode},
% xmajorgrids,
% xmin=-24.3, xmax=532.3,
% xtick style={color=black},
% y grid style={white!69.0196078431373!black},
% ylabel={Reward},
% ymajorgrids,
% ymin=-21.25, ymax=446.25,
% ytick style={color=black}
% ]
\definecolor{color0}{rgb}{0.83921568627451,0.152941176470588,0.156862745098039}
\definecolor{color1}{rgb}{0.12156862745098,0.966666666666667,0.205882352941177}
\definecolor{color2}{rgb}{0.580392156862745,0.503921568627451,0.941176470588235}
\definecolor{color3}{rgb}{0.290196078431372,0.166666666666667,0.23078431372549}

\begin{axis}[
    width=\figurewidth,
    height=\figureheight,
    axis background/.style={fill=white},
    axis line style={white},
    legend cell align={left},
    legend style={fill opacity=0.8, draw opacity=1, text opacity=1, at={(1,0.1)}, anchor=south east, draw=white!80!black},
    tick align=outside,
    tick pos=left,
    x grid style={white},
    xlabel={Episode},
    ymin=-20,
    ymax=420,
    xmajorgrids,
    xtick style={color=black},
    y grid style={white!80!black},
    ylabel={Reward},
    ymajorgrids,
    ytick style={color=black},
    scaled y ticks = false,
    scaled x ticks = false
]
\addplot [semithick, color3, opacity=0.3]
table {%
1 0
2 2
3 2
4 1
5 1
6 5
7 1
8 2
9 1
10 1
11 0
12 3
13 2
14 3
15 3
16 0
17 2
18 2
19 0
20 0
21 1
22 6
23 0
24 1
25 0
26 2
27 0
28 1
29 0
30 5
31 5
32 5
33 5
34 2
35 0
36 4
37 0
38 1
39 0
40 0
41 8
42 8
43 3
44 0
45 3
46 0
47 9
48 1
49 3
50 2
51 3
52 0
53 2
54 3
55 4
56 1
57 1
58 2
59 1
60 4
61 3
62 12
63 3
64 4
65 3
66 6
67 10
68 2
69 2
70 3
71 4
72 4
73 4
74 4
75 6
76 7
77 10
78 7
79 15
80 10
81 7
82 7
83 17
84 8
85 21
86 12
87 9
88 12
89 8
90 9
91 9
92 25
93 20
94 13
95 26
96 4
97 13
98 24
99 29
100 30
101 50
102 78
103 28
104 26
105 65
106 25
107 33
108 34
109 67
110 74
111 30
112 71
113 31
114 19
115 28
116 61
117 12
118 34
119 58
120 73
121 23
122 31
123 70
124 38
125 35
126 24
127 45
128 72
129 68
130 94
131 40
132 48
133 45
134 69
135 96
136 70
137 94
138 32
139 103
140 88
141 48
142 5
143 29
144 36
145 42
146 65
147 47
148 86
149 84
150 39
151 37
152 33
153 102
154 46
155 85
156 62
157 235
158 82
159 50
160 98
161 42
162 66
163 90
164 99
165 86
166 75
167 49
168 40
169 9
170 0
171 64
172 114
173 31
174 49
175 64
176 69
177 90
178 58
179 172
180 49
181 125
182 252
183 31
184 45
185 235
186 168
187 148
188 225
189 121
190 35
191 6
192 88
193 114
194 56
195 203
196 23
197 49
198 39
199 124
200 120
201 62
202 34
203 40
204 51
205 252
206 9
207 43
208 46
209 103
210 126
211 84
212 30
213 36
214 218
215 344
216 320
217 223
218 356
219 103
220 169
221 151
222 60
223 124
224 108
225 162
226 91
227 60
228 94
229 95
230 308
231 304
232 41
233 69
234 332
235 143
236 99
237 288
238 176
239 211
240 93
241 222
242 348
243 262
244 70
245 63
246 229
247 150
248 99
249 76
250 165
251 359
252 81
253 158
254 49
255 174
256 128
257 410
258 112
259 327
260 69
261 340
262 399
263 324
264 30
265 254
266 39
267 412
268 209
269 81
270 306
271 86
272 366
273 363
274 352
275 372
276 119
277 125
278 107
279 135
280 71
281 208
282 394
283 143
284 171
285 310
286 57
287 386
288 111
289 346
290 326
291 362
292 316
293 118
294 336
295 350
296 305
297 106
298 371
299 74
300 1
301 0
302 2
303 12
304 0
305 11
306 128
307 343
308 88
309 81
310 262
311 346
312 334
313 325
314 324
315 378
316 205
317 331
318 392
319 313
320 104
321 371
322 394
323 411
324 369
325 243
326 62
327 115
328 0
329 0
330 11
331 3
332 0
333 307
334 54
335 244
336 43
337 374
338 130
339 58
340 309
341 191
342 346
343 403
344 100
345 355
346 417
347 70
348 217
349 65
350 325
351 396
352 75
353 214
354 195
355 65
356 133
357 360
358 156
359 229
360 46
361 109
362 56
363 3
364 12
365 73
366 97
367 295
368 384
369 394
370 99
371 76
372 155
373 93
374 407
375 242
376 331
377 41
378 0
379 0
380 4
381 48
382 359
383 337
384 407
385 355
386 269
387 317
388 288
389 86
390 243
391 383
392 224
393 346
394 406
395 341
396 328
397 380
398 371
399 285
400 7
401 3
402 261
403 166
404 408
405 138
406 399
407 215
408 423
409 338
410 386
411 398
412 401
413 184
414 327
415 205
416 405
417 378
418 364
419 241
420 335
421 157
422 376
423 124
424 46
425 348
426 89
427 400
428 368
429 414
430 112
431 181
432 276
433 417
434 305
435 218
436 425
437 37
438 373
439 407
440 90
441 183
442 348
443 172
444 194
445 381
446 114
447 120
448 322
449 103
450 423
451 211
452 394
453 415
454 394
455 410
456 93
457 79
458 390
459 181
460 330
461 400
462 384
463 228
464 71
465 149
466 411
467 115
468 129
469 412
470 169
471 390
472 327
473 83
474 273
475 340
476 95
477 18
478 399
479 383
480 56
481 17
482 368
483 400
484 361
485 191
486 97
487 53
488 276
489 295
490 208
491 339
492 175
493 395
494 189
495 115
496 58
497 410
498 238
499 395
500 371
501 59
502 113
503 354
504 229
505 185
506 320
507 378
};
%\addlegendentry{Training}
\end{axis}

\end{tikzpicture}

%% file: fig/breakout/logdir_1B_5m2_p.tex
% This file was created by tikzplotlib v0.9.2.
\begin{tikzpicture}

% \begin{axis}[
% axis background/.style={fill=white!82.7450980392157!black},
% axis line style={white},
% legend cell align={left},
% legend style={fill opacity=0.8, draw opacity=1, text opacity=1, at={(1,0.1)}, anchor=south east, draw=white!80!black},
% tick align=outside,
% tick pos=left,
% x grid style={white!69.0196078431373!black},
% xlabel={Episode},
% xmajorgrids,
% xmin=-76.45, xmax=1627.45,
% xtick style={color=black},
% y grid style={white!69.0196078431373!black},
% ylabel={Reward},
% ymajorgrids,
% ymin=-17.15, ymax=360.15,
% ytick style={color=black}
% ]
\definecolor{color0}{rgb}{0.83921568627451,0.152941176470588,0.156862745098039}
\definecolor{color1}{rgb}{0.12156862745098,0.966666666666667,0.205882352941177}
\definecolor{color2}{rgb}{0.580392156862745,0.503921568627451,0.941176470588235}
\definecolor{color3}{rgb}{0.290196078431372,0.166666666666667,0.23078431372549}

\begin{axis}[
    width=\figurewidth,
    height=\figureheight,
    axis background/.style={fill=white},
    axis line style={white},
    legend cell align={left},
    legend style={fill opacity=0.8, draw opacity=1, text opacity=1, at={(1,0.1)}, anchor=south east, draw=white!80!black},
    tick align=outside,
    tick pos=left,
    x grid style={white},
    xlabel={Episode},
    ymin=-20,
    ymax=420,
    xmajorgrids,
    xtick style={color=black},
    y grid style={white!80!black},
    ylabel={Reward},
    ymajorgrids,
    ytick style={color=black},
    scaled y ticks = false,
    scaled x ticks = false
]
\addplot [semithick, color3, opacity=0.3]
table {%
1 0
2 4
3 2
4 0
5 1
6 1
7 2
8 0
9 2
10 1
11 0
12 3
13 2
14 1
15 3
16 0
17 0
18 1
19 1
20 0
21 1
22 5
23 1
24 0
25 1
26 0
27 0
28 0
29 0
30 5
31 5
32 5
33 4
34 0
35 0
36 4
37 0
38 0
39 0
40 0
41 10
42 4
43 2
44 3
45 1
46 2
47 9
48 1
49 2
50 3
51 0
52 1
53 1
54 0
55 1
56 2
57 1
58 1
59 2
60 6
61 5
62 13
63 2
64 3
65 1
66 1
67 15
68 1
69 13
70 4
71 1
72 3
73 3
74 2
75 2
76 3
77 7
78 15
79 14
80 6
81 7
82 8
83 3
84 13
85 18
86 7
87 14
88 6
89 10
90 6
91 7
92 18
93 8
94 14
95 16
96 12
97 9
98 17
99 15
100 9
101 12
102 19
103 13
104 19
105 22
106 13
107 12
108 20
109 15
110 23
111 10
112 7
113 31
114 19
115 16
116 34
117 6
118 31
119 21
120 18
121 19
122 30
123 28
124 24
125 39
126 28
127 13
128 50
129 26
130 37
131 67
132 17
133 41
134 10
135 10
136 15
137 19
138 11
139 25
140 26
141 42
142 41
143 21
144 59
145 58
146 28
147 51
148 71
149 74
150 49
151 44
152 47
153 33
154 30
155 48
156 61
157 17
158 32
159 91
160 56
161 38
162 60
163 35
164 23
165 41
166 33
167 43
168 64
169 28
170 87
171 76
172 28
173 78
174 46
175 69
176 60
177 31
178 50
179 9
180 68
181 60
182 35
183 100
184 31
185 39
186 24
187 18
188 7
189 87
190 34
191 89
192 50
193 47
194 42
195 36
196 40
197 27
198 23
199 39
200 37
201 20
202 59
203 53
204 11
205 56
206 125
207 54
208 86
209 13
210 25
211 40
212 2
213 81
214 109
215 86
216 88
217 30
218 91
219 31
220 101
221 24
222 0
223 59
224 79
225 98
226 65
227 65
228 35
229 63
230 21
231 34
232 145
233 75
234 27
235 45
236 100
237 86
238 84
239 16
240 33
241 52
242 87
243 19
244 36
245 76
246 27
247 14
248 78
249 44
250 30
251 110
252 262
253 24
254 49
255 23
256 109
257 92
258 53
259 150
260 40
261 47
262 66
263 93
264 114
265 35
266 45
267 104
268 65
269 42
270 99
271 66
272 41
273 46
274 23
275 39
276 224
277 75
278 59
279 40
280 64
281 116
282 50
283 61
284 70
285 84
286 52
287 14
288 26
289 49
290 106
291 112
292 81
293 72
294 70
295 30
296 24
297 78
298 14
299 10
300 37
301 24
302 92
303 33
304 18
305 343
306 94
307 93
308 47
309 66
310 48
311 160
312 27
313 93
314 25
315 46
316 116
317 45
318 19
319 30
320 25
321 34
322 55
323 16
324 78
325 34
326 53
327 37
328 66
329 25
330 0
331 4
332 1
333 49
334 4
335 53
336 38
337 17
338 58
339 56
340 94
341 45
342 32
343 27
344 7
345 49
346 31
347 79
348 267
349 40
350 68
351 242
352 73
353 50
354 73
355 58
356 74
357 78
358 46
359 22
360 33
361 24
362 41
363 73
364 158
365 60
366 29
367 34
368 56
369 61
370 121
371 78
372 47
373 19
374 63
375 92
376 7
377 0
378 61
379 58
380 87
381 66
382 40
383 31
384 85
385 39
386 55
387 62
388 55
389 49
390 4
391 185
392 68
393 11
394 51
395 50
396 49
397 43
398 52
399 55
400 48
401 237
402 35
403 67
404 0
405 47
406 56
407 100
408 30
409 13
410 57
411 70
412 112
413 50
414 55
415 35
416 58
417 18
418 8
419 24
420 3
421 0
422 0
423 1
424 21
425 25
426 20
427 71
428 29
429 45
430 9
431 64
432 37
433 49
434 32
435 57
436 21
437 31
438 48
439 33
440 50
441 73
442 58
443 69
444 50
445 27
446 46
447 23
448 27
449 35
450 137
451 39
452 102
453 40
454 41
455 57
456 37
457 72
458 34
459 21
460 49
461 40
462 45
463 86
464 10
465 93
466 39
467 2
468 2
469 12
470 32
471 92
472 49
473 73
474 26
475 63
476 3
477 22
478 99
479 80
480 51
481 57
482 47
483 61
484 71
485 33
486 44
487 37
488 39
489 24
490 89
491 40
492 23
493 39
494 39
495 55
496 9
497 46
498 30
499 37
500 36
501 100
502 58
503 45
504 28
505 59
506 38
507 25
508 18
509 6
510 6
511 9
512 2
513 8
514 5
515 11
516 93
517 37
518 59
519 36
520 35
521 104
522 10
523 48
524 59
525 78
526 33
527 25
528 20
529 39
530 55
531 53
532 16
533 40
534 19
535 35
536 31
537 69
538 20
539 17
540 14
541 2
542 11
543 29
544 22
545 44
546 23
547 42
548 48
549 9
550 31
551 0
552 6
553 6
554 2
555 0
556 0
557 33
558 10
559 34
560 14
561 10
562 7
563 29
564 14
565 4
566 4
567 10
568 5
569 4
570 13
571 9
572 7
573 4
574 6
575 16
576 12
577 9
578 0
579 7
580 4
581 0
582 7
583 1
584 7
585 0
586 2
587 0
588 25
589 5
590 19
591 1
592 1
593 2
594 0
595 0
596 0
597 0
598 7
599 0
600 7
601 0
602 1
603 0
604 0
605 1
606 1
607 0
608 0
609 0
610 0
611 0
612 0
613 7
614 0
615 0
616 1
617 0
618 1
619 1
620 11
621 1
622 11
623 1
624 11
625 3
626 1
627 0
628 0
629 3
630 0
631 0
632 0
633 0
634 0
635 0
636 0
637 0
638 0
639 0
640 1
641 0
642 11
643 0
644 0
645 0
646 1
647 3
648 2
649 0
650 7
651 2
652 0
653 0
654 1
655 8
656 6
657 1
658 4
659 0
660 0
661 3
662 0
663 2
664 0
665 8
666 3
667 0
668 0
669 16
670 2
671 0
672 1
673 8
674 0
675 3
676 1
677 0
678 7
679 2
680 0
681 0
682 5
683 3
684 8
685 2
686 3
687 2
688 1
689 2
690 5
691 1
692 3
693 3
694 0
695 0
696 0
697 7
698 0
699 0
700 3
701 0
702 0
703 7
704 3
705 22
706 2
707 8
708 3
709 3
710 0
711 0
712 1
713 2
714 8
715 2
716 3
717 2
718 1
719 2
720 2
721 2
722 3
723 11
724 2
725 8
726 4
727 2
728 2
729 0
730 2
731 2
732 2
733 2
734 2
735 2
736 2
737 2
738 2
739 4
740 0
741 3
742 2
743 4
744 3
745 2
746 3
747 2
748 2
749 2
750 2
751 2
752 2
753 2
754 2
755 2
756 2
757 3
758 2
759 2
760 3
761 2
762 2
763 2
764 4
765 0
766 2
767 2
768 3
769 2
770 2
771 2
772 2
773 2
774 3
775 3
776 2
777 2
778 2
779 2
780 2
781 2
782 2
783 3
784 5
785 2
786 6
787 7
788 0
789 0
790 2
791 2
792 2
793 2
794 2
795 2
796 2
797 3
798 4
799 2
800 12
801 3
802 1
803 1
804 3
805 5
806 2
807 0
808 3
809 1
810 2
811 0
812 0
813 3
814 0
815 6
816 0
817 3
818 0
819 1
820 0
821 0
822 0
823 2
824 1
825 0
826 1
827 1
828 3
829 3
830 1
831 2
832 0
833 1
834 1
835 0
836 1
837 1
838 0
839 1
840 0
841 0
842 0
843 0
844 4
845 0
846 1
847 0
848 0
849 0
850 0
851 0
852 0
853 1
854 0
855 5
856 2
857 0
858 0
859 2
860 2
861 1
862 0
863 0
864 2
865 2
866 7
867 2
868 1
869 0
870 0
871 1
872 0
873 1
874 2
875 0
876 3
877 0
878 0
879 0
880 0
881 0
882 0
883 1
884 0
885 0
886 0
887 0
888 1
889 2
890 0
891 0
892 1
893 0
894 0
895 0
896 4
897 0
898 0
899 0
900 0
901 9
902 0
903 0
904 0
905 2
906 1
907 3
908 1
909 1
910 1
911 0
912 0
913 3
914 1
915 2
916 1
917 1
918 0
919 2
920 0
921 2
922 3
923 4
924 3
925 0
926 0
927 1
928 0
929 0
930 0
931 0
932 1
933 1
934 1
935 0
936 3
937 2
938 3
939 2
940 3
941 0
942 1
943 0
944 11
945 0
946 0
947 0
948 0
949 0
950 0
951 11
952 0
953 11
954 11
955 0
956 11
957 0
958 0
959 0
960 0
961 0
962 0
963 0
964 0
965 1
966 1
967 11
968 0
969 2
970 0
971 2
972 6
973 4
974 0
975 3
976 0
977 3
978 2
979 1
980 0
981 0
982 0
983 0
984 2
985 1
986 1
987 0
988 0
989 1
990 0
991 3
992 1
993 0
994 1
995 0
996 0
997 0
998 3
999 0
1000 3
1001 3
1002 1
1003 0
1004 3
1005 0
1006 0
1007 1
1008 3
1009 2
1010 0
1011 0
1012 0
1013 2
1014 0
1015 0
1016 0
1017 1
1018 0
1019 0
1020 0
1021 0
1022 0
1023 0
1024 2
1025 1
1026 1
1027 0
1028 0
1029 0
1030 0
1031 0
1032 0
1033 1
1034 0
1035 7
1036 2
1037 2
1038 0
1039 0
1040 0
1041 0
1042 0
1043 0
1044 0
1045 1
1046 3
1047 0
1048 0
1049 0
1050 1
1051 1
1052 0
1053 4
1054 0
1055 3
1056 5
1057 0
1058 2
1059 2
1060 2
1061 2
1062 2
1063 2
1064 2
1065 2
1066 2
1067 3
1068 2
1069 1
1070 0
1071 0
1072 0
1073 0
1074 1
1075 0
1076 0
1077 0
1078 3
1079 2
1080 2
1081 1
1082 0
1083 0
1084 0
1085 0
1086 3
1087 1
1088 0
1089 2
1090 1
1091 0
1092 0
1093 0
1094 7
1095 0
1096 7
1097 1
1098 0
1099 0
1100 3
1101 1
1102 1
1103 0
1104 1
1105 0
1106 5
1107 3
1108 1
1109 0
1110 0
1111 0
1112 0
1113 0
1114 0
1115 2
1116 0
1117 0
1118 3
1119 2
1120 0
1121 0
1122 4
1123 0
1124 2
1125 0
1126 0
1127 0
1128 2
1129 0
1130 8
1131 0
1132 6
1133 5
1134 1
1135 0
1136 7
1137 14
1138 2
1139 8
1140 4
1141 6
1142 0
1143 0
1144 1
1145 0
1146 0
1147 2
1148 10
1149 0
1150 2
1151 0
1152 0
1153 1
1154 0
1155 0
1156 2
1157 2
1158 0
1159 4
1160 0
1161 0
1162 0
1163 0
1164 0
1165 0
1166 0
1167 0
1168 0
1169 2
1170 7
1171 0
1172 0
1173 1
1174 1
1175 3
1176 0
1177 0
1178 2
1179 2
1180 0
1181 7
1182 2
1183 0
1184 7
1185 1
1186 1
1187 0
1188 1
1189 3
1190 0
1191 7
1192 1
1193 0
1194 1
1195 1
1196 1
1197 0
1198 2
1199 2
1200 0
1201 0
1202 1
1203 0
1204 0
1205 0
1206 0
1207 1
1208 0
1209 0
1210 1
1211 1
1212 1
1213 0
1214 1
1215 1
1216 7
1217 1
1218 0
1219 0
1220 0
1221 0
1222 1
1223 3
1224 0
1225 0
1226 0
1227 1
1228 3
1229 0
1230 2
1231 3
1232 0
1233 0
1234 0
1235 0
1236 2
1237 0
1238 0
1239 0
1240 0
1241 3
1242 0
1243 1
1244 0
1245 1
1246 1
1247 1
1248 0
1249 1
1250 0
1251 0
1252 1
1253 0
1254 2
1255 0
1256 0
1257 1
1258 0
1259 0
1260 1
1261 0
1262 1
1263 1
1264 1
1265 3
1266 3
1267 0
1268 0
1269 0
1270 1
1271 0
1272 2
1273 0
1274 3
1275 1
1276 0
1277 0
1278 0
1279 0
1280 1
1281 3
1282 3
1283 2
1284 3
1285 2
1286 2
1287 1
1288 2
1289 0
1290 1
1291 1
1292 3
1293 2
1294 3
1295 0
1296 0
1297 0
1298 1
1299 0
1300 1
1301 0
1302 7
1303 0
1304 7
1305 0
1306 3
1307 3
1308 0
1309 0
1310 0
1311 3
1312 1
1313 0
1314 0
1315 0
1316 0
1317 2
1318 0
1319 0
1320 1
1321 2
1322 0
1323 0
1324 0
1325 0
1326 1
1327 2
1328 0
1329 0
1330 1
1331 1
1332 0
1333 1
1334 0
1335 0
1336 0
1337 0
1338 1
1339 2
1340 0
1341 0
1342 1
1343 0
1344 0
1345 2
1346 1
1347 0
1348 7
1349 2
1350 1
1351 0
1352 0
1353 1
1354 0
1355 0
1356 3
1357 1
1358 0
1359 1
1360 0
1361 0
1362 0
1363 0
1364 0
1365 1
1366 1
1367 7
1368 1
1369 7
1370 1
1371 1
1372 0
1373 1
1374 0
1375 0
1376 1
1377 0
1378 0
1379 0
1380 0
1381 0
1382 0
1383 0
1384 2
1385 0
1386 0
1387 2
1388 1
1389 3
1390 0
1391 0
1392 1
1393 1
1394 0
1395 1
1396 0
1397 0
1398 0
1399 0
1400 0
1401 3
1402 0
1403 0
1404 1
1405 1
1406 0
1407 1
1408 0
1409 0
1410 1
1411 0
1412 0
1413 0
1414 3
1415 3
1416 1
1417 0
1418 7
1419 1
1420 0
1421 0
1422 2
1423 2
1424 0
1425 7
1426 1
1427 0
1428 6
1429 0
1430 1
1431 1
1432 0
1433 0
1434 2
1435 3
1436 1
1437 0
1438 0
1439 0
1440 0
1441 1
1442 0
1443 0
1444 2
1445 2
1446 0
1447 0
1448 1
1449 0
1450 0
1451 2
1452 0
1453 0
1454 1
1455 0
1456 0
1457 0
1458 0
1459 1
1460 0
1461 1
1462 2
1463 2
1464 0
1465 0
1466 2
1467 2
1468 1
1469 0
1470 0
1471 0
1472 0
1473 1
1474 0
1475 0
1476 0
1477 0
1478 1
1479 0
1480 0
1481 2
1482 2
1483 0
1484 0
1485 1
1486 0
1487 0
1488 2
1489 2
1490 0
1491 7
1492 1
1493 0
1494 0
1495 0
1496 0
1497 0
1498 0
1499 1
1500 2
1501 0
1502 0
1503 1
1504 1
1505 0
1506 1
1507 0
1508 0
1509 0
1510 0
1511 1
1512 1
1513 0
1514 1
1515 3
1516 3
1517 1
1518 0
1519 0
1520 1
1521 0
1522 1
1523 3
1524 1
1525 7
1526 1
1527 3
1528 0
1529 0
1530 1
1531 0
1532 0
1533 1
1534 0
1535 1
1536 1
1537 1
1538 0
1539 1
1540 1
1541 1
1542 4
1543 3
1544 0
1545 1
1546 5
1547 0
1548 3
1549 0
1550 0
};
%\addlegendentry{Training}
\end{axis}

\end{tikzpicture}

%% file: fig/breakout/logdir_2B_10m2_p.tex
% This file was created by tikzplotlib v0.9.2.
\begin{tikzpicture}

% \begin{axis}[
% axis background/.style={fill=white!82.7450980392157!black},
% axis line style={white},
% legend cell align={left},
% legend style={fill opacity=0.8, draw opacity=1, text opacity=1, at={(1,0.1)}, anchor=south east, draw=white!80!black},
% tick align=outside,
% tick pos=left,
% x grid style={white!69.0196078431373!black},
% xlabel={Episode},
% xmajorgrids,
% xmin=-34.45, xmax=745.45,
% xtick style={color=black},
% y grid style={white!69.0196078431373!black},
% ylabel={Reward},
% ymajorgrids,
% ymin=-19.6, ymax=411.6,
% ytick style={color=black}
% ]
\definecolor{color0}{rgb}{0.83921568627451,0.152941176470588,0.156862745098039}
\definecolor{color1}{rgb}{0.12156862745098,0.966666666666667,0.205882352941177}
\definecolor{color2}{rgb}{0.580392156862745,0.503921568627451,0.941176470588235}
\definecolor{color3}{rgb}{0.290196078431372,0.166666666666667,0.23078431372549}

\begin{axis}[
    width=\figurewidth,
    height=\figureheight,
    axis background/.style={fill=white},
    axis line style={white},
    legend cell align={left},
    legend style={fill opacity=0.8, draw opacity=1, text opacity=1, at={(1,0.1)}, anchor=south east, draw=white!80!black},
    tick align=outside,
    tick pos=left,
    x grid style={white},
    xlabel={Episode},
    ymin=-20,
    ymax=420,
    xmajorgrids,
    xtick style={color=black},
    y grid style={white!80!black},
    ylabel={Reward},
    ymajorgrids,
    ytick style={color=black},
    scaled y ticks = false,
    scaled x ticks = false
]
\addplot [semithick, color3, opacity=0.3]
table {%
1 0
2 4
3 0
4 3
5 1
6 2
7 0
8 1
9 0
10 0
11 1
12 0
13 3
14 3
15 2
16 4
17 3
18 1
19 1
20 0
21 0
22 3
23 3
24 2
25 0
26 0
27 0
28 4
29 0
30 9
31 6
32 9
33 10
34 0
35 1
36 9
37 0
38 2
39 2
40 0
41 4
42 4
43 1
44 1
45 3
46 6
47 5
48 1
49 2
50 3
51 1
52 1
53 2
54 7
55 2
56 3
57 1
58 9
59 1
60 6
61 5
62 5
63 7
64 11
65 4
66 4
67 9
68 3
69 4
70 6
71 2
72 6
73 7
74 14
75 3
76 7
77 8
78 8
79 9
80 7
81 11
82 9
83 7
84 10
85 19
86 20
87 7
88 13
89 20
90 13
91 35
92 28
93 26
94 22
95 28
96 28
97 37
98 37
99 6
100 41
101 28
102 31
103 27
104 25
105 61
106 30
107 34
108 42
109 21
110 17
111 58
112 48
113 26
114 17
115 18
116 225
117 49
118 43
119 36
120 12
121 38
122 58
123 29
124 51
125 21
126 73
127 30
128 81
129 47
130 44
131 72
132 47
133 77
134 24
135 47
136 37
137 86
138 42
139 37
140 80
141 23
142 57
143 43
144 55
145 50
146 68
147 78
148 38
149 81
150 41
151 68
152 49
153 29
154 44
155 40
156 136
157 66
158 58
159 63
160 60
161 139
162 31
163 78
164 19
165 38
166 83
167 211
168 139
169 48
170 111
171 18
172 53
173 67
174 120
175 132
176 49
177 62
178 0
179 0
180 31
181 92
182 65
183 48
184 143
185 56
186 238
187 84
188 253
189 77
190 91
191 84
192 59
193 76
194 50
195 60
196 42
197 7
198 26
199 360
200 119
201 10
202 320
203 87
204 123
205 111
206 64
207 120
208 37
209 53
210 337
211 88
212 259
213 54
214 180
215 79
216 102
217 44
218 23
219 104
220 65
221 19
222 30
223 25
224 45
225 30
226 15
227 118
228 207
229 25
230 97
231 42
232 37
233 358
234 115
235 304
236 164
237 318
238 76
239 4
240 20
241 133
242 23
243 36
244 71
245 86
246 14
247 86
248 237
249 74
250 91
251 92
252 45
253 221
254 106
255 53
256 69
257 79
258 137
259 33
260 24
261 103
262 69
263 56
264 73
265 30
266 355
267 31
268 142
269 287
270 26
271 177
272 335
273 77
274 318
275 59
276 85
277 80
278 62
279 98
280 346
281 111
282 41
283 31
284 34
285 84
286 64
287 49
288 83
289 48
290 57
291 42
292 50
293 206
294 36
295 25
296 74
297 40
298 37
299 27
300 300
301 86
302 62
303 79
304 56
305 53
306 10
307 56
308 226
309 156
310 366
311 95
312 116
313 62
314 34
315 62
316 97
317 104
318 47
319 78
320 126
321 61
322 67
323 37
324 62
325 58
326 76
327 32
328 62
329 352
330 105
331 98
332 117
333 58
334 61
335 91
336 176
337 40
338 94
339 91
340 374
341 133
342 25
343 22
344 2
345 0
346 2
347 6
348 99
349 46
350 92
351 28
352 127
353 110
354 94
355 342
356 10
357 57
358 139
359 64
360 58
361 101
362 126
363 117
364 34
365 122
366 13
367 112
368 44
369 21
370 65
371 11
372 24
373 97
374 51
375 63
376 346
377 33
378 50
379 65
380 23
381 27
382 130
383 23
384 276
385 39
386 107
387 40
388 130
389 280
390 43
391 91
392 339
393 56
394 97
395 29
396 370
397 46
398 153
399 20
400 57
401 381
402 12
403 340
404 35
405 18
406 72
407 29
408 22
409 58
410 60
411 377
412 120
413 79
414 132
415 91
416 16
417 20
418 66
419 113
420 31
421 64
422 11
423 4
424 22
425 51
426 45
427 392
428 14
429 22
430 21
431 13
432 11
433 9
434 50
435 86
436 15
437 4
438 28
439 39
440 24
441 10
442 54
443 63
444 304
445 92
446 27
447 50
448 34
449 35
450 149
451 81
452 100
453 24
454 72
455 55
456 3
457 11
458 59
459 182
460 91
461 42
462 52
463 27
464 192
465 92
466 50
467 29
468 22
469 44
470 376
471 178
472 132
473 117
474 42
475 97
476 29
477 204
478 44
479 35
480 342
481 58
482 191
483 38
484 6
485 278
486 16
487 5
488 12
489 1
490 14
491 29
492 11
493 10
494 7
495 6
496 0
497 23
498 76
499 67
500 19
501 77
502 32
503 43
504 5
505 4
506 46
507 41
508 22
509 108
510 230
511 46
512 244
513 385
514 65
515 26
516 53
517 79
518 29
519 19
520 35
521 112
522 90
523 185
524 96
525 134
526 292
527 21
528 11
529 19
530 11
531 7
532 6
533 44
534 42
535 16
536 26
537 22
538 28
539 70
540 28
541 37
542 21
543 44
544 177
545 52
546 34
547 64
548 65
549 36
550 45
551 56
552 36
553 123
554 96
555 71
556 17
557 29
558 47
559 38
560 14
561 69
562 71
563 33
564 44
565 2
566 16
567 10
568 0
569 15
570 7
571 4
572 0
573 24
574 39
575 24
576 22
577 32
578 85
579 34
580 43
581 34
582 36
583 69
584 51
585 33
586 30
587 22
588 29
589 0
590 11
591 36
592 39
593 55
594 76
595 11
596 45
597 18
598 1
599 9
600 0
601 0
602 0
603 0
604 0
605 0
606 0
607 0
608 12
609 6
610 6
611 13
612 22
613 12
614 22
615 12
616 2
617 17
618 4
619 32
620 9
621 50
622 18
623 7
624 19
625 33
626 12
627 32
628 19
629 34
630 10
631 0
632 0
633 9
634 0
635 8
636 28
637 23
638 24
639 4
640 0
641 9
642 0
643 0
644 0
645 9
646 0
647 3
648 9
649 14
650 30
651 57
652 195
653 41
654 42
655 39
656 0
657 36
658 26
659 37
660 47
661 49
662 18
663 9
664 64
665 44
666 40
667 32
668 24
669 21
670 32
671 11
672 30
673 33
674 40
675 24
676 37
677 61
678 60
679 75
680 33
681 61
682 16
683 6
684 15
685 22
686 41
687 35
688 44
689 47
690 0
691 0
692 15
693 25
694 10
695 1
696 15
697 14
698 12
699 18
700 43
701 22
702 49
703 21
704 26
705 21
706 31
707 20
708 35
709 17
710 43
};
%\addlegendentry{Training}
\end{axis}

\end{tikzpicture}

%% file: fig/breakout/logdir_4B_20m2_p.tex
% This file was created by tikzplotlib v0.9.2.
\begin{tikzpicture}

% \begin{axis}[
% axis background/.style={fill=white!82.7450980392157!black},
% axis line style={white},
% legend cell align={left},
% legend style={fill opacity=0.8, draw opacity=1, text opacity=1, at={(1,0.1)}, anchor=south east, draw=white!80!black},
% tick align=outside,
% tick pos=left,
% x grid style={white!69.0196078431373!black},
% xlabel={Episode},
% xmajorgrids,
% xmin=-29.1, xmax=633.1,
% xtick style={color=black},
% y grid style={white!69.0196078431373!black},
% ylabel={Reward},
% ymajorgrids,
% ymin=-20.75, ymax=435.75,
% ytick style={color=black}
% ]
\definecolor{color0}{rgb}{0.83921568627451,0.152941176470588,0.156862745098039}
\definecolor{color1}{rgb}{0.12156862745098,0.966666666666667,0.205882352941177}
\definecolor{color2}{rgb}{0.580392156862745,0.503921568627451,0.941176470588235}
\definecolor{color3}{rgb}{0.290196078431372,0.166666666666667,0.23078431372549}

\begin{axis}[
    width=\figurewidth,
    height=\figureheight,
    axis background/.style={fill=white},
    axis line style={white},
    legend cell align={left},
    legend style={fill opacity=0.8, draw opacity=1, text opacity=1, at={(1,0.1)}, anchor=south east, draw=white!80!black},
    tick align=outside,
    tick pos=left,
    x grid style={white},
    xlabel={Episode},
    ymin=-20,
    ymax=420,
    xmajorgrids,
    xtick style={color=black},
    y grid style={white!80!black},
    ylabel={Reward},
    ymajorgrids,
    ytick style={color=black},
    scaled y ticks = false,
    scaled x ticks = false
]
\addplot [semithick, color3, opacity=0.3]
table {%
1 0
2 2
3 2
4 1
5 1
6 2
7 0
8 0
9 2
10 0
11 0
12 0
13 3
14 6
15 4
16 0
17 3
18 0
19 0
20 0
21 0
22 9
23 0
24 0
25 0
26 0
27 0
28 0
29 1
30 3
31 5
32 5
33 5
34 2
35 0
36 13
37 3
38 1
39 0
40 1
41 4
42 5
43 3
44 2
45 2
46 1
47 5
48 1
49 2
50 3
51 2
52 0
53 0
54 0
55 0
56 2
57 1
58 4
59 0
60 9
61 0
62 5
63 0
64 4
65 3
66 0
67 6
68 2
69 3
70 2
71 5
72 2
73 1
74 0
75 2
76 2
77 0
78 4
79 5
80 1
81 3
82 4
83 2
84 0
85 9
86 6
87 2
88 7
89 1
90 4
91 6
92 11
93 6
94 9
95 10
96 15
97 7
98 13
99 16
100 5
101 10
102 13
103 14
104 8
105 7
106 15
107 10
108 19
109 9
110 29
111 30
112 14
113 24
114 15
115 14
116 15
117 13
118 14
119 14
120 12
121 18
122 15
123 13
124 9
125 26
126 33
127 23
128 11
129 18
130 31
131 19
132 31
133 47
134 23
135 26
136 29
137 20
138 29
139 38
140 20
141 36
142 42
143 40
144 52
145 24
146 44
147 21
148 56
149 58
150 51
151 25
152 30
153 40
154 48
155 59
156 68
157 37
158 39
159 73
160 79
161 53
162 79
163 69
164 56
165 46
166 23
167 95
168 68
169 67
170 35
171 79
172 14
173 29
174 56
175 102
176 58
177 59
178 93
179 30
180 87
181 137
182 90
183 22
184 53
185 10
186 44
187 72
188 68
189 87
190 52
191 58
192 203
193 81
194 78
195 71
196 74
197 122
198 43
199 324
200 52
201 66
202 50
203 221
204 30
205 55
206 220
207 80
208 91
209 2
210 176
211 131
212 36
213 108
214 85
215 109
216 257
217 81
218 194
219 44
220 171
221 72
222 81
223 123
224 134
225 339
226 142
227 116
228 46
229 227
230 121
231 93
232 101
233 240
234 190
235 56
236 168
237 273
238 64
239 373
240 50
241 173
242 82
243 54
244 78
245 167
246 145
247 78
248 407
249 186
250 125
251 82
252 78
253 400
254 47
255 333
256 366
257 217
258 77
259 401
260 372
261 59
262 46
263 272
264 98
265 187
266 395
267 79
268 363
269 394
270 332
271 117
272 81
273 306
274 87
275 32
276 32
277 130
278 134
279 129
280 75
281 77
282 281
283 51
284 141
285 120
286 74
287 290
288 55
289 75
290 372
291 377
292 348
293 45
294 364
295 382
296 109
297 151
298 94
299 110
300 175
301 400
302 144
303 83
304 149
305 149
306 41
307 60
308 277
309 119
310 95
311 386
312 117
313 100
314 281
315 82
316 82
317 40
318 149
319 401
320 194
321 96
322 63
323 79
324 223
325 245
326 63
327 408
328 53
329 77
330 134
331 352
332 45
333 34
334 353
335 403
336 366
337 353
338 97
339 30
340 112
341 262
342 199
343 333
344 58
345 348
346 32
347 56
348 92
349 8
350 39
351 117
352 30
353 43
354 30
355 0
356 93
357 45
358 48
359 406
360 78
361 64
362 378
363 143
364 332
365 110
366 166
367 55
368 83
369 65
370 1
371 1
372 11
373 0
374 23
375 96
376 99
377 42
378 344
379 27
380 205
381 80
382 331
383 242
384 18
385 47
386 29
387 66
388 74
389 355
390 75
391 39
392 227
393 158
394 197
395 236
396 39
397 335
398 110
399 33
400 18
401 15
402 22
403 46
404 155
405 133
406 118
407 12
408 82
409 377
410 85
411 374
412 59
413 36
414 1
415 244
416 70
417 41
418 59
419 145
420 26
421 50
422 55
423 75
424 51
425 372
426 69
427 26
428 339
429 58
430 367
431 147
432 153
433 78
434 59
435 107
436 313
437 118
438 396
439 151
440 33
441 54
442 37
443 119
444 121
445 216
446 30
447 44
448 72
449 152
450 41
451 64
452 80
453 75
454 31
455 122
456 100
457 87
458 74
459 129
460 34
461 39
462 56
463 338
464 321
465 129
466 404
467 139
468 39
469 41
470 195
471 50
472 35
473 36
474 64
475 91
476 367
477 42
478 28
479 77
480 48
481 103
482 47
483 86
484 154
485 63
486 53
487 72
488 110
489 349
490 118
491 137
492 64
493 60
494 127
495 37
496 27
497 77
498 118
499 76
500 57
501 164
502 106
503 51
504 35
505 57
506 179
507 9
508 0
509 9
510 0
511 46
512 95
513 85
514 78
515 31
516 59
517 99
518 43
519 191
520 38
521 104
522 71
523 95
524 94
525 95
526 276
527 97
528 158
529 325
530 40
531 89
532 306
533 38
534 159
535 96
536 47
537 280
538 88
539 107
540 354
541 25
542 25
543 58
544 88
545 21
546 11
547 10
548 90
549 38
550 65
551 44
552 160
553 37
554 126
555 101
556 49
557 7
558 2
559 21
560 30
561 15
562 1
563 18
564 44
565 49
566 79
567 64
568 39
569 39
570 56
571 57
572 81
573 82
574 369
575 31
576 27
577 49
578 17
579 51
580 41
581 18
582 11
583 8
584 352
585 50
586 415
587 68
588 65
589 11
590 15
591 92
592 25
593 139
594 122
595 111
596 66
597 13
598 36
599 90
600 91
601 36
602 48
603 66
};
%\addlegendentry{Training}
\end{axis}

\end{tikzpicture}

%% file: sections/05_Conclusion.tex
\section{Conclusion and Future Work}\label{sec:conclusion}

Adversarial agents and closing the simulation-to-reality gap are among the key challenges preventing wider adoption of reinforcement learning in real-world applications. In this paper, we have addressed the latter one from the perspective of the former: by introducing adversarial conditions inspired by real-world perturbances to a subset of agents in a multi-robot system during a collaborative reinforcement learning process, we have been able to identify points where the robustness of distributed multi-agent DRL algorithms needs to be improved. In this paper, we have considered multiple robotic arms in a simulation environment collaborating towards learning a common policy to reach an object. In order to emulate more realistic conditions and understand how perturbances in the environment affect the learning process, we have considered variability across the agents in terms of their ability to sense and actuate accurately. We have shown how different types of disturbances in the model's input (sensing) and output (actuation) affect the robustness and ability to converge towards an effective policy. We have seen how variable perturbances have the most effect on the ability of the network to converge, while disturbances in the ability of the robots to actuate properly have had a comparatively worse effect than those in their ability to sense the position of the object accurately.

The conclusions of this work serve as a starting point towards the design and development of more robust methods able to identify and take into account these disturbances in the environment that do not occur across all robots equally. This will be the subject of our future work, as well as the study of other types or combinations of disturbances in the environment. We will also work towards modeling more accurately real-world errors for RL simulation environments.